\documentclass[11pt]{article}
\usepackage{jmlr2e}

\usepackage[figuresleft]{rotating}

\usepackage[T1]{fontenc}

\newcommand{\upPsi}{\Psi}
\newcommand{\upGamma}{\Gamma}

\usepackage{natbib}
\usepackage{amsmath}
\usepackage{latexsym}
\usepackage{bbm}
\usepackage{bm}
\usepackage{framed}
\usepackage{epsfig}
\usepackage{url}
\usepackage{microtype}
\usepackage{mdwlist}
\usepackage{algorithm,algorithmic}
\usepackage{bm}

\newcommand{\mysec}[1]{Section~\ref{sec:#1}}

\newcommand{\myeq}[1]{Equation~\ref{eq:#1}}

\newcommand{\myfig}[1]{Figure~\ref{fig:#1}}

\newcommand{\g}{\,\vert\,}
\newcommand{\E}{\mathbb{E}}
\newcommand{\Eq}{\mathbb{E}_q}
\newcommand{\vct}[1]{#1}

\newcommand{\bv}{v}
\newcommand{\mult}{\textrm{Multinomial}}
\newcommand{\dir}{\textrm{Dirichlet}}

\newcommand{\Bet}{\textrm{Beta}}

\newcommand{\bpi}{\mbox{\boldmath$\pi$}}

\newcommand{\bw}{\mbox{\boldmath$w$}}
\newcommand{\bz}{\mbox{\boldmath$z$}}

\newcommand{\cL}{\mathcal{L}}
\newcommand{\N}{\mathcal{N}}

\renewcommand{\bm}[1]{#1}
\newcommand{\myparagraph}[1]{\textit{#1} \,}

\ShortHeadings{Stochastic Variational Inference}{Hoffman, Blei,
 Wang, and Paisley}
\firstpageno{1}

\jmlrheading{}{2013, in press}{}{}{}{Hoffman, Blei, Wang, and Paisley}

\begin{document}

\title{Stochastic Variational Inference}

\author{\name Matt Hoffman \\ 
  \addr{Adobe Research} \\ \email{mathoffm@adobe.com} \AND
  \name David M. Blei \\ \addr{Department of Computer Science} \\
  \addr{Princeton University} \\ \email{blei@cs.princeton.edu} \AND
  \name Chong Wang \\ \addr{Machine Learning Department} \\
  \addr{Carnegie Mellon University} \\ \email{chongw@cs.cmu.edu} \AND
  \name John Paisley \\ \addr{Department of Computer Science} \\
  \addr{University of California, Berkeley} \\
  \email{jpaisley@berkeley.edu} }

\editor{Tommi Jaakkola}

\maketitle

\begin{abstract}
  We develop stochastic variational inference, a scalable algorithm
  for approximating posterior distributions.  We develop this
  technique for a large class of probabilistic models and we
  demonstrate it with two probabilistic topic models, latent Dirichlet
  allocation and the hierarchical Dirichlet process topic model.
  Using stochastic variational inference, we analyze several large
  collections of documents: 300K articles from \textit{Nature}, 1.8M
  articles from \textit{The New York Times}, and 3.8M articles from
  \textit{Wikipedia}.  Stochastic inference can easily handle data
  sets of this size and outperforms traditional variational inference,
  which can only handle a smaller subset.  (We also show that the
  Bayesian nonparametric topic model outperforms its parametric
  counterpart.)  Stochastic variational inference lets us apply
  complex Bayesian models to massive data sets.
\end{abstract}

%% \begin{center}
%%   \textsc{DRAFT: PLEASE DO NOT CITE OR DISTRIBUTE}
%% \end{center}

\section{Introduction \label{sec:intro}}

Modern data analysis requires computation with massive data.  As
examples, consider the following. (1) We have an archive of the raw
text of two million books, scanned and stored online. We want to
discover the themes in the texts, organize the books by subject, and
build a navigator for users to explore our collection. (2) We have
data from an online shopping website containing millions of users'
purchase histories as well as descriptions of each item in the
catalog. We want to recommend items to users based on this
information. (3) We are continuously collecting data from an online
feed of photographs. We want to build a classifier from these
data. (4) We have measured the gene sequences of millions of people.
We want to make hypotheses about connections between observed genes
and other traits.

These problems illustrate some of the challenges to modern data
analysis.  Our data are complex and high-dimensional; we have
assumptions to make---from science, intuition, or other data
analyses---that involve structures we believe exist in the data but
that we cannot directly observe; and finally our data sets are large,
possibly even arriving in a never-ending stream.

Statistical machine learning research has addressed some of these
challenges by developing the field of probabilistic modeling, a field
that provides an elegant approach to developing new methods for
analyzing
data~\citep{Pearl:1988,Jordan:1999a,Bishop:2006,Koller:2009,Murphy:2012}.
In particular, \textit{probabilistic graphical models} give us a
visual language for expressing assumptions about data and its hidden
structure.  The corresponding \textit{posterior inference algorithms}
let us analyze data under those assumptions, inferring the hidden
structure that best explains our observations.

In descriptive tasks, like problems \#1 and \#4 above, graphical
models help us explore the data---the organization of books or the
connections between genes and traits---with the hidden structure
probabilistically ``filled in.'' In predictive tasks, like problems
\#2 and \#3, we use models to form predictions about new observations.
For example, we can make recommendations to users or predict the class
labels of new images.  With graphical models, we enjoy a powerful
suite of probability models to connect and combine; and we have
general-purpose computational strategies for connecting models to data
and estimating the quantities needed to use them.

The problem we face is scale. Inference algorithms of the 1990s and
2000s used to be considered scalable, but they cannot easily handle
the amount of data that we described in the four examples above. This
is the problem we address here. We present an approach to computing
with graphical models that is appropriate for massive data sets, data
that might not fit in memory or even be stored locally. Our method
does not require clusters of computers or specialized hardware, though
it can be further sped up with these amenities.

\begin{figure}
  \begin{center}
    \includegraphics[width=\textwidth]{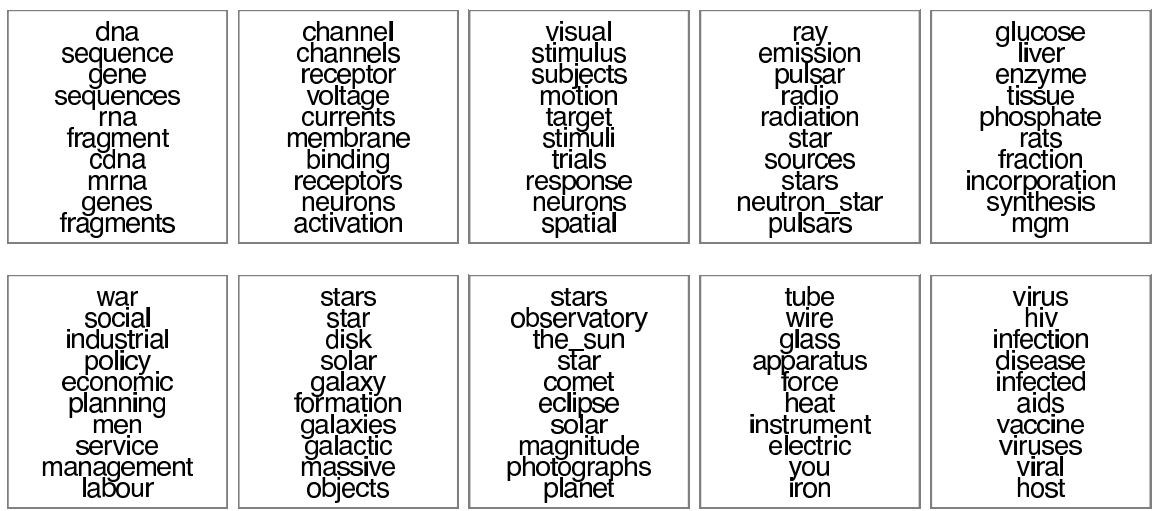}
  \end{center}
  \caption{\label{fig:nyt-topics} Posterior topics from the
    hierarchical Dirichlet process topic model on two large data sets.
    These posteriors were approximated using stochastic variational
    inference with 1.8M articles from the \textit{New York Times}
    (top) and 350K articles from \textit{Nature} (bottom).  (See
    \mysec{hdp} for the modeling details behind the hierarchical
    Dirichlet process and \mysec{experiments} for details about the
    empirical study.)  Each topic is a weighted distribution over the
    vocabulary and each topic's plot illustrates its most frequent
    words.}
\end{figure}

As an example of this approach to data analysis, consider topic
models.  Topic models are probabilistic models of text used to uncover
the hidden thematic structure in a collection of
documents~\citep{Blei:2012}.  The main idea in a topic model is that
there are a set of topics that describe the collection and each
document exhibits those topics with different degrees.  As a
probabilistic model, the topics and how they relate to the documents
are hidden structure and the main computational problem is to infer
this hidden structure from an observed collection.  \myfig{nyt-topics}
illustrates the results of our algorithm on a probabilistic topic
model.  These are two sets of topics, weighted distributions over the
vocabulary, found in 1.8M articles from the \textit{New York Times}
and 300,000 articles from \textit{Nature}.  Topic models are motivated
by applications that require analyzing massive collections of
documents like this, but traditional algorithms for topic model
inference do not easily scale collections of this size.

% For example, in a collection of news articles some of the topics might
% include \textit{business}, \textit{war}, \textit{sports}, and
% \textit{health}; an article, such as one about the health benefits of
% joining casual sports leagues, would combine the \textit{sports} and
% \textit{health} topics.

% Topic models have become a successful case-study in applied
% probabilistic modeling.  They are probabilistic models where the
% topics and per-document topic proportions are hidden variables.  The
% topics are weighted distributions over the vocabulary; the
% per-document proportions are weighted distributions over the topics.
% The central computational problem for topic modeling is posterior
% inference---given an observed collection of documents, our goal is to
% compute the conditional distribution of the topics and topic
% proportions.  They are motivated by applications that require
% analyzing massive collections of documents.  (For example, topic
% models are used in problem \#1 above.)  ~\citep{Blei:2012}

% dmb: below i liked this sentence but it actually doesn't fit.

% Modern algorithms must handle as much data as is available, computing
% not with data sets but with data sources.

Our algorithm builds on variational inference, a method that transforms
complex inference problems into high-dimensional optimization problems
\citep{Jordan:1999,Wainwright:2008}.  Traditionally, the optimization
is solved with a coordinate ascent algorithm, iterating between
re-analyzing every data point in the data set and re-estimating its
hidden structure. This is inefficient for large data sets, however,
because it requires a full pass through the data at each iteration.

In this paper we derive a more efficient algorithm by using stochastic
optimization~\citep{Robbins:1951}, a technique that follows noisy
estimates of the gradient of the objective.  When used in variational
inference, we show that this gives an algorithm which iterates between
subsampling the data and adjusting the hidden structure based only on
the subsample.  This is much more efficient than traditional
variational inference.  We call our method \textit{stochastic
  variational inference}.

We will derive stochastic variational inference for a large class of
graphical models. We will study its performance on two kinds of
probabilistic topic models.  In particular, we demonstrate stochastic
variational inference on latent Dirichlet allocation
\citep{Blei:2003b}, a simple topic model, and the hierarchical
Dirichlet process topic model \citep{Teh:2006b}, a more flexible model
where the number of discovered topics grows with the data. (This
latter application demonstrates how to use stochastic variational
inference in a variety of Bayesian nonparametric settings.) Stochastic
variational inference can efficiently analyze massive data sets with
complex probabilistic models.

\myparagraph{Technical summary.} We now turn to the technical context of
our method. In probabilistic modeling, we use hidden variables to
encode hidden structure in observed data; we articulate the
relationship between the hidden and observed variables with a
factorized probability distribution (i.e., a graphical model); and we
use inference algorithms to estimate the posterior distribution, the
conditional distribution of the hidden structure given the
observations.

Consider a graphical model of hidden and observed random variables for
which we want to compute the posterior. For many models of interest,
this posterior is not tractable to compute and we must appeal to
approximate methods. The two most prominent strategies in statistics
and machine learning are Markov chain Monte Carlo (MCMC) sampling and
variational inference. In MCMC sampling, we construct a Markov chain
over the hidden variables whose stationary distribution is the
posterior of interest
\citep{Metropolis:1953,Hastings:1970,Geman:1984,Gelfand:1990,Robert:2004}. We
run the chain until it has (hopefully) reached equilibrium and collect
samples to approximate the posterior. In variational inference, we
define a flexible family of distributions over the hidden variables,
indexed by free parameters \citep{Jordan:1999,Wainwright:2008}.  We
then find the setting of the parameters (i.e., the member of the
family) that is closest to the posterior. Thus we solve the inference
problem by solving an optimization problem.

Neither MCMC nor variational inference scales easily to the kinds of
settings described in the first paragraph. Researchers have proposed
speed-ups of both approaches, but these usually are tailored to
specific models or compromise the correctness of the algorithm (or
both). Here, we develop a general variational method that scales.

As we mentioned above, the main idea in this work is to use stochastic
optimization \citep{Robbins:1951, Spall:2003}. In stochastic
optimization, we find the maximum of an objective function by
following noisy (but unbiased) estimates of its gradient. 
Under the right conditions, stochastic optimization
%% If the expectation of the
%% noisy gradient equals the true gradient then stochastic optimization
algorithms provably converge to an optimum of the objective.
Stochastic optimization is particularly attractive when the objective
(and therefore its gradient) is a sum of many terms that can be
computed independently. In that setting, we can cheaply compute noisy
gradients by subsampling only a few of these terms.

Variational inference is amenable to stochastic optimization because
the variational objective decomposes into a sum of terms, one for each
data point in the analysis. We can cheaply obtain noisy estimates of
the gradient by subsampling the data and computing a scaled gradient
on the subsample.  If we sample independently then the expectation of
this noisy gradient is equal to the true gradient.  With one more
detail---the idea of a natural
gradient~\citep{Amari:1998}---stochastic variational inference has an
attractive form:
\begin{enumerate*}
\item Subsample one or more data points from the data.
\item Analyze the subsample using the current variational parameters.
\item Implement a closed-form update of the variational parameters.
\item Repeat.
\end{enumerate*}
While traditional algorithms require repeatedly analyzing the whole
data set before updating the variational parameters, this algorithm
only requires that we analyze randomly sampled subsets.  We will show
how to use this algorithm for a large class of graphical models.

\myparagraph{Related work.}  Variational inference for probabilistic
models was pioneered in the mid-1990s.  In Michael Jordan's lab, the
seminal papers of \cite{Saul:1996,Saul:1996a} and \cite{Jaakkola:1997}
grew out of reading the statistical physics
literature~\citep{Peterson:1987,Parisi:1988}.  In parallel, the
mean-field methods explained in~\cite{Neal:1999} (originally published
in 1993) and~\cite{Hinton:1993} led to variational algorithms for
mixtures of experts~\citep{Waterhouse:1996}.

In subsequent years, researchers began to understand the potential for
variational inference in more general settings and developed generic
algorithms for conjugate exponential-family
models~\citep{Attias:1999,Attias:2000,Wiegerinck:2000,Ghahramani:2001,Xing:2003}.
These innovations led to automated variational inference, allowing a
practitioner to write down a model and immediately use variational
inference to estimate its posterior~\citep{Bishop:2003}.  For good
reviews of variational inference see~\cite{Jordan:1999}
and~\cite{Wainwright:2008}.

In this paper, we develop scalable methods for generic Bayesian
inference by solving the variational inference problem with stochastic
optimization~\citep{Robbins:1951}.  Our algorithm builds on the
earlier approach of~\cite{Sato:2001}, whose algorithm only applies to
the limited set of models that can be fit with the EM
algorithm~\citep{Dempster:1977}.  Specifically, we generalize his
approach to the much wider set of probabilistic models that are
amenable to closed-form coordinate ascent inference.  Further, in the
sense that EM itself is a mean-field method~\citep{Neal:1999}, our
algorithm builds on the stochastic optimization approach to
EM~\citep{Cappe:2009}.  Finally, we note that stochastic optimization
was also used with variational inference in~\cite{Platt:2008} for fast
approximate inference in a specific model of web service activity.

For approximate inference, the main alternative to variational methods
is Markov chain Monte Carlo (MCMC)~\citep{Robert:2004}.  Despite its
popularity in Bayesian inference, relatively little work has focused
on developing MCMC algorithms that can scale to very large data sets.
One exception is sequential Monte Carlo, although these typically lack
strong convergence guarantees~\citep{Doucet:2001}. Another is the
stochastic gradient Langevin method of \citet{Welling:2011}, which
enjoys asymptotic convergence guarantees and also takes advantage of
stochastic optimization.  Finally, in topic modeling, researchers have
developed several approaches to parallel
MCMC~\citep{Newman:2009a,Smola:2010,Ahmed:2012}.

\myparagraph{The organization of this paper.} In \mysec{inference}, we
review variational inference for graphical models and then derive
stochastic variational inference. In \mysec{topic-models}, we review
probabilistic topic models and Bayesian nonparametric models and then
derive the stochastic variational inference algorithms in these
settings. In \mysec{experiments}, we study stochastic variational
inference on several large text data sets.

\section{Stochastic Variational Inference \label{sec:inference}}

We derive \textit{stochastic variational inference}, a stochastic
optimization algorithm for mean-field variational inference.  Our
algorithm approximates the posterior distribution of a probabilistic
model with hidden variables, and can handle massive data sets of
observations.

We divide this section into four parts.
\begin{enumerate}
\item We define the class of models to which our algorithm applies.
  We define \textit{local} and \textit{global} hidden variables, and
  requirements on the conditional distributions within the model.

\item We review \textit{mean-field variational inference}, an
  approximate inference strategy that seeks a tractable distribution
  over the hidden variables which is close to the posterior
  distribution.  We derive the traditional variational inference
  algorithm for our class of models, which is a coordinate ascent
  algorithm.

\item We review the \textit{natural gradient} and derive the natural
  gradient of the variational objective function.  The natural
  gradient closely relates to coordinate ascent variational inference.

\item We review stochastic optimization, a technique that uses noisy
  estimates of a gradient to optimize an objective function, and apply
  it to variational inference.  Specifically, we use stochastic
  optimization with noisy estimates of the natural gradient of the
  variational objective.  These estimates arise from repeatedly
  subsampling the data set.  We show how the resulting algorithm,
  \textit{stochastic variational inference}, easily builds on
  traditional variational inference algorithms but can handle much
  larger data sets.

\end{enumerate}

\subsection{Models with local and global hidden variables}

\begin{figure}
  \begin{center}
    \includegraphics[width=0.4\textwidth]{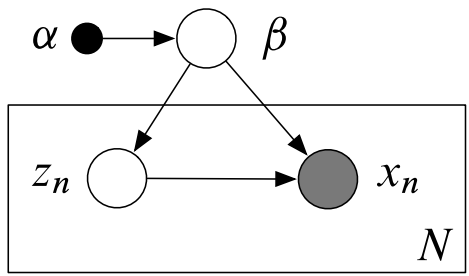}
  \end{center}
  \caption{A graphical model with observations $x_{1:N}$, local hidden
    variables $z_{1:N}$ and global hidden variables $\beta$.  The
    distribution of each observation $x_n$ only depends on its
    corresponding local variable $z_n$ and the global variables
    $\beta$.  (Though not pictured, each hidden variable $z_n$,
    observation $x_n$, and global variable $\beta$ may be a collection
    of multiple random variables.) \label{fig:gm}}
\end{figure}

Our class of models involves observations, global hidden variables,
local hidden variables, and fixed parameters.  The $N$ observations
are $x = x_{1:N}$; the vector of global hidden variables is $\beta$;
the $N$ local hidden variables are $z = z_{1:N}$, each of which is a
collection of $J$ variables $z_n = z_{n,1:J}$; the vector of fixed
parameters is $\alpha$.  (Note we can easily allow $\alpha$ to partly
govern any of the random variables, such as fixed parts of the
conditional distribution of observations. To keep notation simple, we
assume that they only govern the global hidden variables.)

The joint distribution factorizes into a global term and a product of
local terms,
\begin{equation}
  \label{eq:joint}
  p(x, z, \beta \g \alpha) =
  p(\beta \g \alpha) \prod_{n=1}^{N} p(x_n, z_n \g
  \beta).
\end{equation}
\myfig{gm} illustrates the graphical model.  Our goal is to
approximate the posterior distribution of the hidden variables given
the observations, $p(\beta, z \g x)$.

The distinction between local and global hidden variables is
determined by the conditional dependencies.  In particular, the $n$th
observation $x_n$ and the $n$th local variable $z_n$ are conditionally
independent, given global variables $\beta$, of all other observations
and local hidden variables,
\begin{equation*}
  p(x_n, z_n \g x_{-n}, z_{-n}, \beta, \alpha) = p(x_n, z_n \g \beta,
  \alpha).
\end{equation*}
The notation $x_{-n}$ and $z_{-n}$ refers to the set of variables
except the $n$th.

This kind of model frequently arises in Bayesian statistics. The
global variables $\beta$ are parameters endowed with a prior
$p(\beta)$ and each local variable $z_n$ contains the hidden structure
that governs the $n$th observation.  For example, consider a Bayesian
mixture of Gaussians.  The global variables are the mixture
proportions and the means and variances of the mixture components; the
local variable $z_n$ is the hidden cluster label for the $n$th
observation $x_n$.

We have described the independence assumptions of the hidden
variables.  We make further assumptions about the \textit{complete
  conditionals} in the model.  A complete conditional is the
conditional distribution of a hidden variable given the other hidden
variables and the observations.  We assume that these distributions
are in the exponential family,
\begin{align}
  \label{eq:global-cond}
  p(\beta \g x, z, \alpha) &= h(\beta) \exp\{\eta_{g}(x, z,
  \alpha)^\top
  t(\beta) - a_g(\eta_{g}(x, z, \alpha))\} \\
  \label{eq:local-cond}
  p(z_{nj} \g x_n, z_{n,-j}, \beta) &= h(z_{nj})
  \exp\{\eta_{\ell}(x_n, z_{n,-j}, \beta)^{\top} t(z_{nj}) -
  a_{\ell}(\eta_{\ell}(x_n, z_{n,-j}, \beta))\}.
\end{align}
The scalar functions $h(\cdot)$ and $a(\cdot)$ are respectively the
{\it base measure} and {\it log-normalizer}; the vector functions
$\eta(\cdot)$ and $t(\cdot)$ are respectively the {\it natural
  parameter} and {\it sufficient statistics}.\footnote{We use
  overloaded notation for the functions $h(\cdot)$ and $t(\cdot)$ so
  that they depend on the names of their arguments; for example,
  $h(z_{nj})$ can be thought of as a shorthand for the more formal
  (but more cluttered) notation $h_{z_{nj}}(z_{nj})$. This is
  analogous to the standard convention of overloading the probability
  function $p(\cdot)$.} These are conditional distributions, so the
natural parameter is a function of the variables that are being
conditioned on.  (The subscripts on the natural parameter $\eta$
indicate complete conditionals for local or global variables.)  For
the local variables $z_{nj}$, the complete conditional distribution
is determined by the global variables $\beta$ and the other local
variables in the $n$th context, i.e., the $n$th data point $x_n$ and
the local variables $z_{n,-j}$.  This follows from the factorization
in \myeq{joint}.

% dmb: removed "g(\beta)" for brevity.

These assumptions on the complete conditionals imply a conjugacy
relationship between the global variables $\beta$ and the local
contexts $(z_n, x_n)$, and this relationship implies a specific form
of the complete conditional for $\beta$.  Specifically, the
distribution of the local context given the global variables must be in an
exponential family,
\begin{equation}
\label{eq:xzprior}
p(x_n, z_n | \beta) = h(x_n, z_n)\exp\{\beta^\top t(x_n, z_n)
- a_{\ell}(\beta)\}.
\end{equation}
The prior distribution $p(\beta)$ must also be in an exponential
family,
\begin{equation}
\label{eq:betaprior}
p(\beta) = h(\beta)\exp\{\alpha^\top t(\beta) - a_g(\alpha)\}.
\end{equation}
The sufficient statistics are $t(\beta) = (\beta,
-a_\ell(\beta))$ and thus the hyperparameter $\alpha$ has two
components $\alpha = ( \alpha_1, \alpha_2 )$.  The first component
$\alpha_1$ is a vector of the same dimension as $\beta$; the second
component $\alpha_2$ is a scalar.

Equations \ref{eq:xzprior} and \ref{eq:betaprior} imply that the
complete conditional for the global variable in \myeq{global-cond} is
in the same exponential family as the prior with natural parameter
\begin{equation}
  \label{eq:global-posterior}
  \eta_g(x, z, \alpha) = (\alpha_1 + \textstyle \sum_{n=1}^{N}
  t(z_n, x_n), \alpha_2 + N).
\end{equation}
This form will be important when we derive stochastic variational
inference in \mysec{stochastic-variational-inference}.  See
\citet{Bernardo:1994} for a general discussion of conjugacy and the
exponential family.

This family of distributions---those with local and global variables,
and where the complete conditionals are in the exponential
family---contains many useful statistical models from the machine
learning and statistics literature.  Examples include Bayesian mixture
models \citep{Ghahramani:2000, Attias:2000}, latent Dirichlet
allocation~\citep{Blei:2003b}, hidden Markov models (and many
variants) \citep{Rabiner:1989,Fine:1998,Fox:2011b, Paisley:2009},
Kalman filters (and many variants) \citep{Kalman:1960, Fox:2011a},
factorial models~\citep{Ghahramani:1997}, hierarchical linear
regression models~\citep{Gelman:2007}, hierarchical probit
classification models~\citep{McCullagh:1989, Girolami:2006},
probabilistic factor analysis/matrix factorization
models~\citep{Spearman:1904,Tipping:1999,Collins:2002,Wang:2006,Salakhutdinov:2008,
  Paisley:2009, Hoffman:2010}, certain Bayesian nonparametric mixture
models~\citep{Antoniak:1974,Escobar:1995,Teh:2006b}, and
others.\footnote{We note that our assumptions can be relaxed to the
  case where the full conditional $p(\beta|x, z)$ is not tractable,
  but each partial conditional $p(\beta_k | x, z, \beta_{-k})$
  associated with the global variable $\beta_k$ is in a tractable
  exponential family.  The topic models of the next section do not
  require this complexity, so we chose to keep the derivation a little
  simpler.}

% !!! DMB: once written, point to the appendix in the footnote above.

Analyzing data with one of these models amounts to computing the
posterior distribution of the hidden variables given the observations,
\begin{equation}
  \label{eq:posterior}
  p(z, \beta \g x) = \frac{p(x, z, \beta)}{\int
    p(x, z, \beta)dz d\beta}.
\end{equation}
We then use this posterior to explore the hidden structure of our data
or to make predictions about future data.  For many models however,
such as the examples listed above, the denominator in \myeq{posterior}
is intractable to compute.  Thus we resort to approximate posterior
inference, a problem that has been a focus of modern Bayesian
statistics.  We now turn to mean-field variational inference, the
approximation inference technique which roots our strategy for
scalable inference.

\subsection{Mean-field variational inference}

Variational inference casts the inference problem as an
optimization.  We introduce a family of distributions over the hidden
variables that is indexed by a set of free parameters, and then
optimize those parameters to find the member of the family that is
closest to the posterior of interest. (Closeness is measured with
Kullback-Leibler divergence.) We use the resulting distribution,
called the \textit{variational distribution}, to approximate the
posterior.

In this section we review mean-field variational inference, the form
of variational inference that uses a family where each hidden variable
is independent.  We describe the variational objective function,
discuss the mean-field variational family, and derive the traditional
coordinate ascent algorithm for fitting the variational parameters.
This algorithm is a stepping stone to stochastic variational
inference.

\myparagraph{The evidence lower bound.} Variational inference minimizes
the Kullback-Leibler (KL) divergence from the variational distribution
to the posterior distribution.  It maximizes the \textit{evidence
  lower bound} (ELBO), a lower bound on the logarithm of the marginal
probability of the observations $\log p(x)$. The ELBO is equal to the
negative KL divergence up to an additive constant.

We derive the ELBO by introducing a distribution over the hidden
variables $q(z, \beta)$ and using Jensen's inequality. (Jensen's
inequality and the concavity of the logarithm function imply that
$\log \E[f(y)] \ge \E[\log f(y)]$ for any random variable $y$.)  This
gives the following bound on the log marginal,
\begin{align}
  \log p(x) &= \log \int p(x, z, \beta) dz d\beta \nonumber \\
  &= \log \int p(x, z, \beta)
  \frac{q(z, \beta)}{q(z, \beta)} dz d\beta \nonumber \\
  &= \log \left(\E_q\left[\frac{p(x, z, \beta)}
      {q(z, \beta)}\right]\right) \nonumber \\
  \label {eq:elbo} &\geq \E_q[\log p(x, z, \beta)]
  - \E_q[\log q(z, \beta)] \\
  &\triangleq {\cal L}(q). \nonumber
\end{align}
The ELBO contains two terms.  The first term is the expected log
joint, $\E_q[\log p(x, z, \beta)]$.  The second term is the entropy of
the variational distribution, $-\E_q[\log q(z, \beta)]$.  Both of
these terms depend on $q(z, \beta)$, the variational distribution of
the hidden variables.

We restrict $q(z, \beta)$ to be in a family that is tractable, one for
which the expectations in the ELBO can be efficiently computed.  We
then try to find the member of the family that maximizes the ELBO.
Finally, we use the optimized distribution as a proxy for the
posterior.

Solving this maximization problem is equivalent to finding the member
of the family that is closest in KL divergence to the
posterior~\citep{Jordan:1999,Wainwright:2008},
\begin{eqnarray*}
\textstyle
\textrm{KL}(q(z, \beta)||p(z, \beta | x))
&=& \Eq\left[\log q(z, \beta)\right] - \Eq\left[\log p(z, \beta \g
  x)\right] \\
&=& \Eq\left[\log q(z, \beta)\right] - \Eq\left[\log p(x, z,
  \beta)\right] + \log p(x) \\
&=& -\cL(q) + \mathrm{const.}
\end{eqnarray*}
$\log p(\bm{x})$ is replaced by a constant because it does not depend
on $q$.

% dmb: changed "convenient" back to simplest.  its the simplest one
% that's actually used.  (that simpler ones are possible is not
% relevant.)

\myparagraph{The mean-field variational family.}  The simplest
variational family of distributions is the \textit{mean-field family}.
In this family, each hidden variable is independent and governed by
its own parameter,
\begin{equation}
  \label{eq:meanfield}
  q(\bm{z}, \beta) = q(\beta \g \lambda)
  \prod_{n=1}^{N} \prod_{j=1}^{J} q(z_{nj} \g \phi_{nj}).
\end{equation}
The global parameters $\lambda$ govern the global variables; the local
parameters $\phi_n$ govern the local variables in the $n$th context.
The ELBO is a function of these parameters.

\myeq{meanfield} gives the factorization of the variational family,
but does not specify its form.  We set $q(\beta|\lambda)$ and
$q(z_{nj}|\phi_{nj})$ to be in the same exponential family as the
complete conditional distributions $p(\beta|x, z)$ and $p(z_{nj}|x_n,
z_{n,-j}, \beta)$, from Equations~\ref{eq:global-cond} and
\ref{eq:local-cond}.  The variational parameters $\lambda$ and
$\phi_{nj}$ are the natural parameters to those families,
\begin{align}
  \label{eq:q-global}
  q(\beta \g \lambda) &=
  h(\beta) \exp\{\lambda^\top t(\beta) - a_g(\lambda)\} \\
  \label{eq:q-local}
  q(z_{nj} \g \phi_{nj}) &=
  h(z_{nj}) \exp\{\phi_{nj}^\top t(z_{nj}) - a_{\ell}(\phi_{nj})\}.
\end{align}
These forms of the variational distributions lead to an easy
coordinate ascent algorithm.  Further, the optimal mean-field
distribution, without regard to its particular functional form, has
factors in these families~\citep{Bishop:2006}.

Note that assuming that these exponential families are the same as
their corresponding conditionals means that $t(\cdot)$ and $h(\cdot)$
in \myeq{q-global} are the same functions as $t(\cdot)$ and $h(\cdot)$
in \myeq{global-cond}.  Likewise, $t(\cdot)$ and $h(\cdot)$ in
\myeq{q-local} are the same as in \myeq{local-cond}. We will sometimes
suppress the explicit dependence on $\phi$ and $\lambda$, substituting
$q(z_{nj})$ for $q(z_{nj}|\phi_{nj})$ and $q(\beta)$ for
$q(\beta|\lambda)$.

The mean-field family has several computational advantages. For one,
the entropy term decomposes,
\begin{equation*}
  - \E_q[\log q(\bm{z}, \beta)] =
  - \E_{\lambda}[\log q(\beta)] -
  \sum_{n=1}^{N} \sum_{j=1}^{J} \E_{\phi_{nj}}[\log q(z_{nj})],
\end{equation*}
where $\E_{\phi_{nj}}[\cdot]$ denotes an expectation with respect to
$q(z_{nj} \g \phi_{nj})$ and $\E_{\lambda}[\cdot]$ denotes an
expectation with respect to $q(\beta \g \lambda)$.  Its other
computational advantages will emerge as we derive the gradients of the
variational objective and the coordinate ascent algorithm.

\myparagraph{The gradient of the ELBO and coordinate ascent inference.}
We have defined the objective function in \myeq{elbo} and the
variational family in Equations~\ref{eq:meanfield}, \ref{eq:q-global}
and \ref{eq:q-local}.  Our goal is to optimize the objective with
respect to the variational parameters.

In traditional mean-field variational inference, we optimize
\myeq{elbo} with coordinate ascent.  We iteratively optimize each
variational parameter, holding the other parameters fixed.  With the
assumptions that we have made about the model and variational
distribution---that each conditional is in an exponential family and
that the corresponding variational distribution is in the same
exponential family---we can optimize each coordinate in closed form.

We first derive the coordinate update for the parameter $\lambda$ to the
variational distribution of the global variables $q(\beta \g
\lambda)$.  As a function of $\lambda$, we can rewrite the objective
as
\begin{equation}
  \label{eq:elbo-lambda}
  {\cal L}(\lambda) = \E_q[\log p(\beta \g x, z)]
  - \E_q[\log q(\beta)] + \textrm{const.}
\end{equation}
The first two terms are expectations that involve $\beta$; the third
term is constant with respect to $\lambda$.  The constant absorbs
quantities that depend only on the other hidden variables.  Those
quantities do not depend on $q(\beta \g \lambda)$ because all
variables are independent in the mean-field family.

\myeq{elbo-lambda} reproduces the full ELBO in \myeq{elbo}.  The
second term of \myeq{elbo-lambda} is the entropy of the global
variational distribution.  The first term derives from the expected
log joint likelihood, where we use the chain rule to separate terms
that depend on the variable $\beta$ from terms that do not,
\begin{equation*}
  \E_q[\log p(x, z, \beta)] = \E_q[\log p(x, z)] + \E_q[\log p(\beta
  \g x, z)].\end{equation*}
The constant absorbs $\E_q[\log p(\bm{x},\bm{z})]$, leaving
the expected log conditional $\E_q[\log p(\beta \g x, z)]$.

Finally, we substitute the form of $q(\beta \g \lambda)$ in
\myeq{q-global} to obtain the final expression for the ELBO as a
function of $\lambda$,
\begin{equation}
  \label{eq:lambdaelbo}
  {\cal L}(\lambda) = \E_q[\eta_g(\bm{x}, \bm{z}, \alpha)]^\top \nabla_\lambda a_g(\lambda)
  - \lambda^\top \nabla_\lambda a_g(\lambda) + a_g(\lambda) + \textrm{const}.
\end{equation}
In the first and second terms on the right side, we used the
exponential family identity that the expectation of the sufficient
statistics is the gradient of the log normalizer,
$\E_{q}[t(\beta)] = \nabla_\lambda a_g(\lambda)$.  The constant
has further absorbed the expected log normalizer of the conditional
distribution $-\E_q[a_g(\eta_g(\bm{x}, \bm{z}, \alpha))]$, which does
not depend on $q(\beta)$.

\myeq{lambdaelbo} simplifies the ELBO as a function of the global
variational parameter.  To derive the coordinate ascent update, we
take the gradient,
\begin{equation}
  \label{eq:global-grad}
  \nabla_{\lambda}{\cal L} =
  \nabla^2_{\lambda} a_g(\lambda) (\E_{q}[\eta_g(\bm{x}, \bm{z},
  \alpha)] - \lambda).
\end{equation}
We can set this gradient to zero by setting
\begin{equation}
  \label{eq:coord-global}
  \lambda = \E_{q}[\eta_g(\bm{x}, \bm{z}, \alpha)].
\end{equation}
This sets the global variational parameter equal to the expected
natural parameter of its complete conditional distribution.
Implementing this update, holding all other variational parameters
fixed, optimizes the ELBO over $\lambda$.  Notice that the mean-field
assumption plays an important role.  The update is the expected
conditional parameter $\E_{q}[\eta_g(\bm{x}, \bm{z}, \alpha)]$,
which is an expectation of a function of the other random variables
and observations.  Thanks to the mean-field assumption, this
expectation is only a function of the local variational parameters and
does not depend on $\lambda$.

We now turn to the local parameters $\phi_{nj}$.  The gradient is
nearly identical to the global case,
\begin{equation*}
  \nabla_{\phi_{nj}} {\cal L} =
  \nabla^2_{\phi_{nj}} a_{\ell}(\phi_{nj})
  (\E_{q}[\eta_{\ell}(x_n, z_{n,-j}, \beta)] - \phi_{nj}).
\end{equation*}
It equals zero when
\begin{equation} \label{eq:coord-local}
  \phi_{nj} = \E_{q}[\eta_\ell(x_n, z_{n,-j}, \beta)].
\end{equation}
Mirroring the global update, this expectation does not depend on
$\phi_{nj}$.  However, while the global update in \myeq{coord-global}
depends on all the local variational parameters---and note there is a
set of local parameters for each of the $N$ observations---the local
update in \myeq{coord-local} only depends on the global parameters and
the other parameters associated with the $n$th context.  The
computational difference between local and global updates will be
important in the scalable algorithm of \mysec{stochastic-variational-inference}.

\begin{figure}
\begin{framed}
\begin{algorithmic}[1]
  \STATE Initialize $\lambda^{(0)}$ randomly.\\
  \REPEAT
  \FOR{each local variational parameter $\phi_{nj}$}
  \STATE Update $\phi_{nj}$,
   $\phi_{nj}^{(t)} =
   \E_{q^{(t-1)}}[\eta_{\ell,j}(x_n, z_{n,-j},\beta)]$.
  \ENDFOR
  \STATE Update the global variational parameters, $\lambda^{(t)} =
  \E_{q^{(t)}}[\eta_{g}(z_{1:N}, x_{1:N})]$.  \UNTIL{the ELBO
    converges}
\end{algorithmic}
\end{framed}
\caption{Coordinate ascent mean-field variational
  inference. \label{fig:classical-vi}}
\end{figure}

The updates in Equations~\ref{eq:coord-global} and
\ref{eq:coord-local} form the algorithm for coordinate ascent
variational inference, iterating between updating each local parameter
and the global parameters.  The full algorithm is in
\myfig{classical-vi}, which is guaranteed to find a local optimum of
the ELBO.  Computing the expectations at each step is easy for
directed graphical models with tractable complete conditionals, and in
\mysec{topic-models} we show that these updates are tractable for
many topic models.  \myfig{classical-vi} is the ``classical''
variational inference algorithm, used in many settings.

% !!! Matt TODO: show the easy computation more rigorously, in an
% appendix.

% DMB: i took this out for pace: "The details use the relationship
% between the sufficient statistics and log-normalizers."  (yes, that
% is "pace" and not "space.")

%% MDH: On further reflection, I'm not sure this paragraph (which
%% reviewer 2 didn't like) is necessary.

% DMB: i disagree.  this paragraph could be a relief for stasticians.
% R2 was overly harsh about it.  and, i further disagree with him that
% this looks like we "invented" the connection.

As an aside, these updates reveal a connection between mean-field
variational inference and Gibbs sampling~\citep{Gelfand:1990}.  In
Gibbs sampling, we iteratively sample from each complete conditional.
In variational inference, we take variational expectations of the
natural parameters of the same distributions. The updates also show a
connection to the expectation-maximization (EM) algorithm
\citep{Dempster:1977}---\myeq{coord-local} corresponds to the E step,
and \myeq{coord-global} corresponds to the M step \citep{Neal:1999}.

We mentioned that the local steps (Steps 3 and 4 in
\myfig{classical-vi}) only require computation with the global
parameters and the $n$th local context.  Thus, the data can be
distributed across many machines and the local variational updates can
be implemented in parallel.  These results can then be aggregated in
Step 6 to find the new global variational parameters.

However, the local steps also reveal an inefficiency in the algorithm.
The algorithm begins by initializing the global parameters $\lambda$
randomly---the initial value of $\lambda$ does not reflect any
regularity in the data. But before completing even one iteration, the
algorithm must analyze every data point using these initial (random)
values.  This is wasteful, especially if we expect that we can learn
something about the global variational parameters from only a subset
of the data.

% DMB: removed this.  we need to handle streaming stuff in a "future
% problems" way, which it is.

% Further, if the data are ``infinite'', i.e., if they represent a data
% source where information arrives in a constant stream, then this
% algorithm can never complete even one iteration.

% DMB: below, i described the algorithm here.  might as well, b/c all
% the pieces are actually in place to implement it.  (i do this when i
% speak about SVI as well.)

We solve this problem with stochastic optimization.  This leads to
stochastic variational inference, an efficient algorithm that
continually improves its estimate of the global parameters as it
analyzes more observations. Though the derivation requires some
details, we have now described all of the computational components of
the algorithm.  (See \myfig{stoch-vi}.)  At each iteration, we sample
a data point from the data set and compute its optimal local
variational parameters; we form \textit{intermediate global
  parameters} using classical coordinate ascent updates where the
sampled data point is repeated $N$ times; finally, we set the new
global parameters to a weighted average of the old estimate and the
intermediate parameters.

The algorithm is efficient because it need not analyze the whole data
set before improving the global variational parameters, and the
per-iteration steps only require computation about a single local
context.  Furthermore, it only uses calculations from classical
coordinate inference.  Any existing implementation of variational
inference can be easily configured to this scalable alternative.

We now show how stochastic inference arises by applying stochastic
optimization to the natural gradients of the variational objective.
We first discuss natural gradients and their relationship to the
coordinate updates in mean-field variational inference.

\subsection{The natural gradient of the ELBO}

The natural gradient of a function accounts for the information
geometry of its parameter space, using a Riemannian metric to adjust
the direction of the traditional gradient.  \citet{Amari:1998}
discusses natural gradients for maximum-likelihood estimation, which
give faster convergence than standard gradients.  In this section we
describe Riemannian metrics for probability distributions and the
natural gradient of the ELBO.

\myparagraph{Gradients and probability distributions.}  The classical
gradient method for maximization tries to find a maximum of a function
$f(\lambda)$ by taking steps of size $\rho$ in the direction of the
gradient,
\begin{equation*}
\lambda^{(t+1)} = \lambda^{(t)} + \rho\nabla_\lambda f(\lambda^{(t)}).
\end{equation*}
The gradient (when it exists) points in the direction of steepest
ascent.  That is, the gradient $\nabla_\lambda f(\lambda)$ points in
the same direction as the solution to
\begin{align}
\label{eq:steepestascent}
\arg\max_{d\lambda} f(\lambda + d\lambda)
\quad \textrm{subject to }||d\lambda||^2 < \epsilon^2
\end{align}
for sufficiently small $\epsilon$.  \myeq{steepestascent} implies that
if we could only move a tiny distance $\epsilon$ away from $\lambda$
then we should move in the direction of the gradient. Initially this
seems reasonable, but there is a complication.  The gradient direction
implicitly depends on the Euclidean distance metric associated with
the space in which $\lambda$ lives. However, the Euclidean metric
might not capture a meaningful notion of distance between settings of
$\lambda$.

% dmb: i edited and then removed this piece (below).  i thought it was
% clearer to be slightly mysterious at the end of the last paragraph
% and then immediately show the reader what we mean by going back to
% our context.

% For example, when $\lambda$ parameterizes a probability
% distribution and the objective is a function of the distribution, the
% Euclidean metric is sensitive to how $\lambda$ might be rescaled
% within the probability density.

% We will focus on the variational distribution of the global variable
% $q(\beta|\lambda)$.

The problem with Euclidean distance is especially clear in our
setting, where we are trying to optimize an objective with respect to
a parameterized probability distribution $q(\beta \g \lambda)$.  When
optimizing over a probability distribution, the Euclidean distance
between two parameter vectors $\lambda$ and $\lambda'$ is often a poor
measure of the dissimilarity of the distributions $q(\beta \g
\lambda)$ and $q(\beta \g \lambda')$.  For example, suppose $q(\beta)$
is a univariate normal and $\lambda$ is the mean $\mu$ and scale
$\sigma$.  The distributions $\N(0, 10000)$ and $\N(10, 10000)$ are
almost indistinguishable, and the Euclidean distance between their
parameter vectors is 10.  In contrast, the distributions $\N(0, 0.01)$
and $\N(0.1, 0.01)$ barely overlap, but this is not reflected in the
Euclidean distance between their parameter vectors, which is only 0.1.
The \textit{natural gradient} corrects for this issue by redefining
the basic definition of the gradient~\citep{Amari:1998}.

\myparagraph{Natural gradients and probability distributions.}  A
natural measure of dissimilarity between probability distributions is
the symmetrized KL divergence
\begin{equation}
\label{eq:kld}
\textstyle
% MDH: I know it seems like there should be a 1/2, but it comes out
% cleaner this way.
D^\mathrm{sym}_{KL}(\lambda, \lambda') =
\E_\lambda \left[\log \frac{q(\beta\g \lambda)}{q(\beta\g \lambda')}\right]
+\E_{\lambda'}\left[\log \frac{q(\beta\g \lambda')}{q(\beta\g \lambda)}\right].
\end{equation}
Symmetrized KL depends on the distributions themselves, rather than on
how they are parameterized; it is invariant to parameter transformations.

With distances defined using symmetrized KL, we find the direction of
steepest ascent in the same way as for gradient methods,
\begin{equation}
\label{eq:naturalascent}
\arg\max_{d\lambda} f(\lambda + d\lambda)
\quad \textrm{subject to} \ D^\mathrm{sym}_{KL}(\lambda, \lambda+d\lambda) < \epsilon.
\end{equation}
As $\epsilon\rightarrow 0$, the solution to this problem points in the
same direction as the \textit{natural gradient}.  While the Euclidean
gradient points in the direction of steepest ascent in Euclidean
space, the natural gradient points in the direction of steepest ascent
in the Riemannian space, i.e., the space where local distance is
defined by KL divergence rather than the $L^2$ norm.

We manage the more complicated constraint in \myeq{naturalascent} with
a Riemannian metric $G(\lambda)$ \citep{Docarmo:1992}.  This metric defines linear
transformations of $\lambda$ under which the squared Euclidean
distance between $\lambda$ and a nearby vector $\lambda+d\lambda$ is
the KL between $q(\beta|\lambda)$ and $q(\beta|\lambda+d\lambda)$,
\begin{equation}
\label{eq:innerproduct}
  d\lambda^T G(\lambda) d\lambda = D^\mathrm{sym}_{KL}(\lambda,
  \lambda+d\lambda),
\end{equation}
and note that the transformation can be a function of $\lambda$.
\citet{Amari:1998} showed that we can compute the natural gradient by
premultiplying the gradient by the inverse of the Riemannian metric
$G(\lambda)^{-1}$,
\begin{equation*}
  \hat\nabla_\lambda f(\lambda) \triangleq G(\lambda)^{-1}\nabla_\lambda
  f(\lambda),
\end{equation*}
where $G$ is the Fisher information matrix of $q(\lambda)$
\citep{Amari:1982,Kullback:1951},
\begin{equation}
\label{eq:fisher}
  G(\lambda) = \E_\lambda\left[(\nabla_{\lambda} \log q(\beta \g \lambda))
    (\nabla_{\lambda} \log q(\beta \g \lambda))^\top\right].
\end{equation}
We can show that \myeq{fisher} satisfies \myeq{innerproduct} by
approximating $\log q(\beta \g \lambda+d\lambda)$ using the
first-order Taylor approximations about $\lambda$
\begin{equation*}
\begin{split}
\log q(\beta | \lambda + d\lambda) &= O(d\lambda^2) + \log q(\beta | \lambda) + d\lambda^\top \nabla_\lambda \log q(\beta | \lambda),
\\ q(\beta | \lambda + d\lambda) &= O(d\lambda^2) + q(\beta | \lambda) + q(\beta | \lambda) d\lambda^\top \nabla_\lambda \log q(\beta | \lambda),
\end{split}
\end{equation*}
and plugging the result into \myeq{kld}:
\begin{equation*}
\begin{split}
D^\mathrm{sym}_{KL}(\lambda, \lambda+d\lambda)
&= \int_\beta (q(\beta | \lambda + d\lambda) - q(\beta | \lambda))
 (\log q(\beta | \lambda + d\lambda) - \log q(\beta | \lambda)) d\beta
\\ &= O(d\lambda^3) + \int_\beta q(\beta | \lambda) (d\lambda^\top \nabla_\lambda \log q(\beta | \lambda))^2 d\beta
\\ &= O(d\lambda^3) + \E_q[(d\lambda^\top \nabla_\lambda \log q(\beta | \lambda))^2]
= O(d\lambda^3) + d\lambda^\top G(\lambda) d\lambda.
\end{split}
\end{equation*}
For small enough $d\lambda$ we can ignore the $O(d\lambda^3)$ term.

When $q(\beta \g \lambda)$ is in the exponential family
(\myeq{q-global}) the metric is the second derivative of the log
normalizer,
\begin{equation*}
\begin{split}
%\label{eq:prob-metric}
G(\lambda) &= \E_\lambda\left[(\nabla_{\lambda} \log p(\beta \g \lambda))
  (\nabla_{\lambda} \log p(\beta \g \lambda))^\top\right] \\
&= \E_\lambda\left[(t(\beta) - \E_\lambda[t(\beta)])
  (t(\beta) - \E_\lambda[t(\beta)])^\top\right] \\
&= \nabla^2_{\lambda} a_g(\lambda).
\end{split}
\end{equation*}
This follows from the exponential family identity that the Hessian of
the log normalizer function $a$ with respect to the natural parameter
$\lambda$ is the covariance matrix of the sufficient statistic vector
$t(\beta)$.

\myparagraph{Natural gradients and mean field variational inference.}
We now return to variational inference and compute the natural
gradient of the ELBO with respect to the variational parameters.
Researchers have used the natural gradient in variational inference
for nonlinear state space models~\citep{Honkela:2008} and Bayesian
mixtures~\citep{Sato:2001}.\footnote{Our work here---using the natural
  gradient in a stochastic optimization algorithm---is closest to that
  of \citet{Sato:2001}, though we develop the algorithm via a
  different path and Sato does not address models for which the joint
  conditional $p(z_{n} | \beta, x_n)$ is not tractable.}

Consider the global variational parameter $\lambda$.  The gradient of
the ELBO with respect to $\lambda$ is in \myeq{global-grad}.  Since
$\lambda$ is a natural parameter to an exponential family
distribution, the Fisher metric defined by $q(\beta)$ is
$\nabla^2_\lambda a_g(\lambda)$.  Note that the Fisher metric is the
first term in \myeq{global-grad}.  We premultiply the gradient by the
inverse Fisher information to find the natural gradient.  This reveals
that the natural gradient has the following simple form,
\begin{equation}
  \label{eq:natgrad}
  \hat{\nabla}_{\lambda} {\cal L} =
  \E_{\phi}[\eta_g(\bm{x}, \bm{z}, \alpha)] - \lambda.
\end{equation}
An analogous computation goes through for the local variational
parameters,
\begin{equation*}
%  \label{eq:natgrad-local}
  \hat{\nabla}_{\phi_{nj}} {\cal L} =
  \E_{\lambda,\phi_{n,-j}}[\eta_\ell(x_n, z_{n, -j}, \beta)] -
  \phi_{nj}.
\end{equation*}
The natural gradients are closely related to the coordinate ascent
updates of \myeq{coord-global} or \myeq{coord-local}.  Consider a full
set of variational parameters $\lambda$ and $\bm{\phi}$.  We can
compute the natural gradient by computing the coordinate updates in
parallel and subtracting the current setting of the parameters.  The
classical coordinate ascent algorithm can thus be interpreted as a
projected natural gradient algorithm~\citep{Sato:2001}.  Updating a
parameter by taking a natural gradient step of length one is
equivalent to performing a coordinate update.

We motivated natural gradients by mathematical reasoning around the
geometry of the parameter space.  More importantly, however, natural
gradients are easier to compute than classical gradients.  They are
easier to compute because premultiplying by the Fisher information
matrix---which we must do to compute the classical gradient
in~\myeq{global-grad} but which disappears from the natural gradient
in \myeq{natgrad}---is prohibitively expensive for variational
parameters with many components.  In the next section we will see that
efficiently computing the natural gradient lets us develop scalable
variational inference algorithms.
\subsection{Stochastic variational inference}
\label{sec:stochastic-variational-inference}

The coordinate ascent algorithm in \myfig{classical-vi} is inefficient
for large data sets because we must optimize the local variational
parameters for each data point before re-estimating the global
variational parameters.  Stochastic variational inference uses
stochastic optimization to fit the global variational parameters.  We
repeatedly subsample the data to form noisy estimates of the natural
gradient of the ELBO, and we follow these estimates with a decreasing
step-size.

We have reviewed mean-field variational inference in models with
exponential family conditionals and showed that the natural gradient
of the variational objective function is easy to compute.  We now
discuss stochastic optimization, which uses a series of noisy
estimates of the gradient, and use it with noisy natural gradients to
derive stochastic variational inference.

\myparagraph{Stochastic optimization.}  Stochastic optimization
algorithms follow noisy estimates of the gradient with a decreasing
step size.  Noisy estimates of a gradient are often cheaper to compute
than the true gradient, and following such estimates can allow
algorithms to escape shallow local optima of complex objective
functions.  In statistical estimation problems, including variational
inference of the global parameters, the gradient can be written as a
sum of terms (one for each data point) and we can compute a fast noisy
approximation by subsampling the data.  With certain conditions on the
step-size schedule, these algorithms provably converge to an
optimum~\citep{Robbins:1951}. \citet{Spall:2003} gives an overview of
stochastic optimization; ~\citet{Bottou:2003} gives an overview of its
role in machine learning.

Consider an objective function $f(\lambda)$ and a random function
$B(\lambda)$ that has expectation equal to the gradient so that
$\E_q[B(\lambda)] = \nabla_\lambda f(\lambda)$.  The stochastic
gradient algorithm, which is a type of stochastic optimization,
optimizes $f(\lambda)$ by iteratively following realizations of
$B(\lambda)$.  At iteration $t$, the update for $\lambda$ is
\begin{equation*}
  \lambda^{(t)} = \lambda^{(t-1)} + \rho_t b_t(\lambda^{(t-1)}),
\end{equation*}
where $b_t$ is an independent draw from the noisy gradient $B$.  If
the sequence of step sizes $\rho_t$ satisfies
\begin{equation}
\begin{split}
  \textstyle \sum \rho_t &= \infty \\
  \textstyle \sum \rho_t^2 &< \infty \label{eq:rho}
\end{split}
\end{equation}
then $\lambda^{(t)}$ will converge to the optimal $\lambda^*$ (if $f$
is convex) or a local optimum of $f$ (if not convex).\footnote{To find
  a local optimum, $f$ must be three-times differentiable and meet a
  few mild technical requirements~\citep{Bottou:1998}.  The
  variational objective satisfies these criteria.}  The same results
apply if we premultiply the noisy gradients $b_t$ by a sequence of
positive-definite matrices $G_t^{-1}$ (whose eigenvalues are bounded)
\citep{Bottou:1998}.  The resulting algorithm is
\begin{equation*}
  \lambda^{(t)} = \lambda^{(t-1)} + \rho_t G_t^{-1} b_t(\lambda^{(t-1)}).
\end{equation*}
As our notation suggests, we will use the Fisher metric for $G_t$,
replacing stochastic Euclidean gradients with stochastic natural
gradients.

\begin{figure}
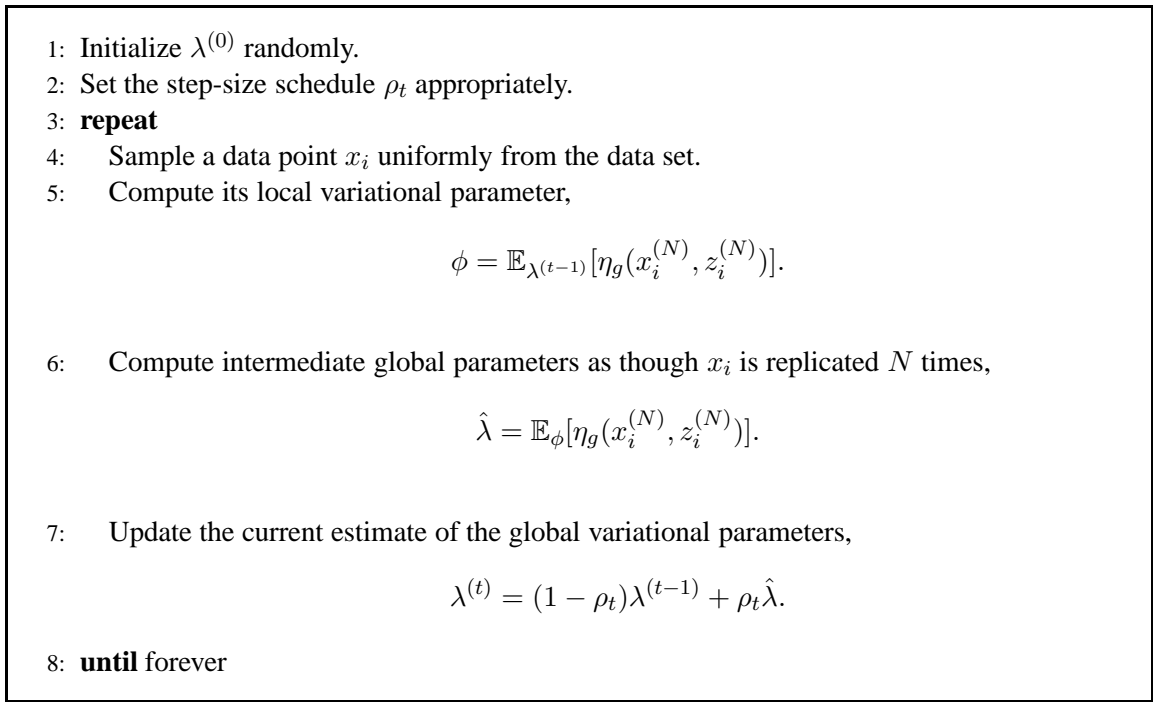

\begin{framed}
\begin{algorithmic}[1]
  \STATE Initialize $\lambda^{(0)}$ randomly.
  \STATE Set the step-size schedule $\rho_t$ appropriately.
  \REPEAT
  \STATE Sample a data point $x_i$ uniformly from the data set.
  \STATE Compute its local variational parameter, \[\phi =
  \E_{\lambda^{(t-1)}}[\eta_{g}(x_i^{(N)}, z_i^{(N)})].\]
  \STATE Compute intermediate global
  parameters as though $x_i$ is replicated $N$ times,
  \[
    \hat{\lambda} =
    \E_{\phi}[\eta_{g}(x_i^{(N)}, z_i^{(N)})].
    \]
  \STATE Update the current estimate
  of the global variational parameters, \[\lambda^{(t)} = (1-\rho_t)
  \lambda^{(t-1)} + \rho_t \hat{\lambda}.\]
  \UNTIL{forever}
\end{algorithmic}
\end{framed}
\caption{Stochastic variational inference. \label{fig:stoch-vi}}
\end{figure}

\myparagraph{Stochastic variational inference.}  We use stochastic
optimization with noisy natural gradients to optimize the variational
objective function.  The resulting algorithm is in
\myfig{stoch-vi}. At each iteration we have a current setting of the
global variational parameters.  We repeat the following steps:
\begin{enumerate}
\item Sample a data point from the set; optimize its local variational
  parameters.
\item Form intermediate global variational parameters, as though we were
  running classical coordinate ascent and the sampled data point were
  repeated $N$ times to form the collection.
\item Update the global variational parameters to be a weighted
  average of the intermediate parameters and their current setting.
\end{enumerate}
We show that this algorithm is stochastic natural gradient ascent on
the global variational parameters.

Our goal is to find a setting of the global variational parameters
$\lambda$ that maximizes the ELBO. Writing $\mathcal{L}$ as a function
of the global and local variational parameters, Let the function
$\phi(\lambda)$ return a local optimum of the local variational
parameters so that
\begin{equation*}
  \nabla_\phi {\cal L}(\lambda, \phi(\lambda)) = 0.
\end{equation*}
Define the \textit{locally maximized ELBO} ${\cal L}(\lambda)$ to be
the ELBO when $\lambda$ is held fixed and the local variational
parameters $\phi$ are set to a local optimum $\phi(\lambda)$,
\begin{equation*}
  % \label{eq:Lhat}
  {\cal L}(\lambda)\triangleq
  {\cal L}(\lambda, \phi(\lambda)).
\end{equation*}
We can compute the (natural) gradient of $\cL(\lambda)$ by first
finding the corresponding optimal local parameters $\phi(\lambda)$ and
then computing the (natural) gradient of ${\cal L}(\lambda,
\phi(\lambda))$, holding $\phi(\lambda)$ fixed.  The reason is that
the gradient of $\cL(\lambda)$ is the same as the gradient of the
two-parameter ELBO ${\cal L}(\lambda, \phi(\lambda))$,
\begin{eqnarray}
\label{eq:Lhatgrad}
\nabla_\lambda{\cal L}(\lambda)
&=& \nabla_\lambda {\cal L}(\lambda, \phi(\lambda))
+ (\nabla_\lambda \phi(\lambda))^\top
\nabla_{\phi} {\cal L}(\lambda, \phi(\lambda)) \\
&=& \nabla_\lambda {\cal L}(\lambda, \phi(\lambda)),
\end{eqnarray}
where $\nabla_\lambda \phi(\lambda)$ is the Jacobian of
$\phi(\lambda)$ and we use the fact that the gradient of ${\cal
  L}(\lambda, \phi)$ with respect to $\phi$ is zero at
$\phi(\lambda)$.

Stochastic variational inference optimizes the maximized ELBO ${\cal
  L}(\lambda)$ by subsampling the data to form noisy estimates of the
natural gradient.  First, we decompose ${\cal
  L}(\lambda)$ into a global term and a sum of local terms,
\begin{equation}
  \label{eq:decomposed-elbo}
  {\cal L}(\lambda) = \E_q[\log p(\beta)] - \E_q[\log q(\beta)] +
  \sum_{n=1}^{N}
  \max_{\phi_n} (\E_q[\log p(x_n, z_n \g \beta)]
  - \E_q[\log q(z_n)]).
\end{equation}

Now consider a variable that chooses an index of the data uniformly at
random, $I \sim \textrm{Unif}(1,\ldots,N)$.  Define ${\cal
  L}_I(\lambda)$ to be the following random function of the
variational parameters,
\begin{equation}
  \label{eq:noisy-elbo}
  {\cal L}_I(\lambda) \triangleq
  \E_q[\log p(\beta)] -
  \E_q[\log q(\beta)]
  +
  N \max_{\phi_I}\left(
    \E_q[\log p(x_I, z_I \g \beta)
    - \E_q[\log q(z_I)]
  \right).
\end{equation}
The expectation of ${\cal L}_I$ is equal to the objective in
\myeq{decomposed-elbo}.  Therefore, the natural gradient of ${\cal
  L}_I$ with respect to each global variational parameter $\lambda$
is a noisy but unbiased estimate of the natural gradient of the
variational objective.  This process---sampling a data point and then
computing the natural gradient of ${\cal L}_I$---will provide cheaply
computed noisy gradients for stochastic optimization.

We now compute the noisy gradient.  Suppose we have sampled the $i$th
data point.  Notice that \myeq{noisy-elbo} is equivalent to the full
objective of \myeq{decomposed-elbo} where the $i$th data point is
observed $N$ times.  Thus the natural gradient of
\myeq{noisy-elbo}---which is a noisy natural gradient of the
ELBO---can be found using \myeq{natgrad},
\begin{equation*}
  \hat{\nabla}{\cal L}_i = \E_{q}\left[\eta_g\left(x_i^{(N)},
  z_i^{(N)}, \alpha\right)\right]-\lambda,
\end{equation*}
where $\left\{x_i^{(N)}, z_i^{(N)}\right\}$ are a data set formed by $N$
replicates of observation $x_n$ and hidden variables $z_n$.

We compute this expression in more detail.  Recall the complete
conditional $\eta_g(\bm{x}, \bm{z}, \alpha)$ from \myeq{global-posterior}.
From this equation, we can compute the conditional natural parameter
for the global parameter given $N$ replicates of $x_n$,
\begin{equation*}
  \eta_g\!\left(x_i^{(N)}, z_i^{(N)}, \alpha\right) =
  \alpha + N \cdot ( t(x_n, z_n), 1).
\end{equation*}
Using this in the natural gradient of \myeq{natgrad} gives a noisy
natural gradient,
\begin{equation*}
  % \label{eq:noisy-natgrad}
  \hat{\nabla}_{\lambda} {\cal L}_i =
  \alpha + N \cdot \left(\E_{\phi_i(\lambda)}[t(x_i, z_i)], 1\right) - \lambda,
\end{equation*}
where $\phi_i(\lambda)$ gives the elements of $\phi(\lambda)$
associated with the $i$th local context.  While the full natural
gradient would use the local variational parameters for the whole data
set, the noisy natural gradient only considers the local parameters
for one randomly sampled data point.  These noisy gradients are
cheaper to compute.

% DMB: i expanded this a bit.  i defined the intermediate parameter
% here to make the notation of the equations cleaner.

Finally, we use the noisy natural gradients in a Robbins-Monro
algorithm to optimize the ELBO.  We sample a data point $x_i$ at each
iteration.  Define the intermediate global parameter $\hat{\lambda}_t$
to be the estimate of $\lambda$ that we would obtain if the sampled
data point was replicated $N$ times,
\begin{equation*}
  \hat{\lambda}_t \triangleq \alpha + N\E_{\phi_i(\lambda)}[( t(x_i,
  z_i), 1 )].
\end{equation*}
This comprises the first two terms of the noisy natural gradient.  At
each iteration we use the noisy gradient (with step size $\rho_t$) to
update the global variational parameter.  The update is
\begin{eqnarray*}
  \lambda^{(t)} &=& \lambda^{(t-1)} + \rho_t \left(\hat{\lambda}_t -
  \lambda^{(t-1)}\right) \\
  &=& (1-\rho_t) \lambda^{(t-1)} +
  \rho_t \hat{\lambda}_t.
\end{eqnarray*}
This is a weighted average of the previous estimate of $\lambda$ and
the estimate of $\lambda$ that we would obtain if the sampled data
point was replicated $N$ times.

\myfig{stoch-vi} presents the full algorithm.  At each iteration, the
algorithm has an estimate of the global variational parameter
$\lambda^{(t-1)}$.  It samples a single data point from the data and
cheaply computes the intermediate global parameter $\hat{\lambda}_t$,
i.e., the next value of $\lambda$ if the data set contained $N$
replicates of the sampled point.  It then sets the new estimate of the
global parameter to be a weighted average of the previous estimate and
the intermediate parameter.

We set the step-size at iteration $t$ as follows,
\begin{equation}
  \label{eq:stepsize}
  \rho_t = (t + \tau)^{-\kappa}.
\end{equation}
This satisfies the conditions in \myeq{rho}. The \textit{forgetting
  rate} $\kappa \in (0.5, 1]$ controls how quickly old information is
forgotten; the \textit{delay} $\tau \geq 0$ down-weights early
iterations. In \mysec{experiments} we fix the delay to be one and
explore a variety of forgetting rates.  Note that this is just one way
to parameterize the learning rate.  As long as the step size
conditions in \myeq{rho} are satisfied, this iterative algorithm
converges to a local optimum of the ELBO.

% !!! Matt TODO: enumerate the conditions from Leon's 2003 article and
% show that the ELBO meets them.  (not needed for submission)

% DMB: it might be okay, in a footnote, to simply say that the ELBO
% meets the conditions of bottou 2003.

\subsection{Extensions}\label{sec:extensions}

We now describe two extensions of the basic stochastic inference
algorithm in \myfig{stoch-vi}: the use of multiple samples
(``minibatches'') to improve the algorithm's stability, and empirical
Bayes methods for hyperparameter estimation.

\myparagraph{Minibatches.}  So far, we have considered stochastic
variational inference algorithms where only one observation $x_t$ is
sampled at a time. Many stochastic optimization algorithms benefit
from ``minibatches,'' i.e., several examples at a
time~\citep{Bottou:2008, Liang:2009, Mairal:2010}. In stochastic
variational inference, we can sample a set of $S$ examples at each
iteration $x_{t,1:S}$ (with or without replacement), compute the local
variational parameters $\phi_s(\lambda^{(t-1)})$ for each data point,
compute the intermediate global parameters $\hat\lambda_s$ for each
data point $x_{ts}$, and finally average the $\hat\lambda_s$ variables
in the update
\begin{equation*}
  \lambda^{(t)} = (1-\rho_t)\lambda^{(t-1)} +
  \frac{\rho_t}{S}\sum_s \hat\lambda_s.
\end{equation*}
The stochastic natural gradients associated with each point $x_s$ have
expected value equal to the gradient.  Therefore, the average of these
stochastic natural gradients has the same expectation and the
algorithm remains valid.

There are two reasons to use minibatches. The first reason is to
amortize any computational expenses associated with updating the
global parameters across more data points; for example, if the
expected sufficient statistics of $\beta$ are expensive to compute,
using minibatches allows us to incur that expense less frequently.
The second reason is that it may help the algorithm to find better
local optima. Stochastic variational inference is guaranteed to
converge to a local optimum but taking large steps on the basis of
very few data points may lead to a poor one. As we will see in
\mysec{experiments}, using more of the data per update can help the
algorithm.

\myparagraph{Empirical Bayes estimation of hyperparameters.}  In some
cases we may want to both estimate the posterior of the hidden random
variables $\beta$ and $z$ and obtain a point estimate of the values of
the hyperparameters $\alpha$.  One approach to fitting $\alpha$ is to
try to maximize the marginal likelihood of the data $p(x \g \alpha)$,
which is also known as empirical Bayes~\citep{Maritz:1989} estimation.
Since we cannot compute $p(x\g\alpha)$ exactly, an approximate
approach is to maximize the fitted variational lower bound ${\cal L}$
over $\alpha$.  In the non-stochastic setting, $\alpha$ can be
optimized by interleaving the coordinate ascent updates in
\myfig{classical-vi} with an update for $\alpha$ that increases the
ELBO.  This is called variational expectation-maximization.

In the stochastic setting, we update $\alpha$ simultaneously with
$\lambda$.  We can take a step in the direction of the gradient of the
noisy ELBO ${\cal L}_t$ (\myeq{noisy-elbo}) with respect to $\alpha$,
scaled by the step-size $\rho_t$,
\begin{equation*}
\alpha^{(t)} = \alpha^{(t-1)} + \rho_t \nabla_\alpha {\cal
  L}_t(\lambda^{(t-1)}, \phi, \alpha^{(t-1)}).
\end{equation*}
Here $\lambda^{(t-1)}$ are the global parameters from the previous
iteration and $\phi$ are the optimized local parameters for the
currently sampled data point. We can also replace the standard
Euclidean gradient with a natural gradient or Newton step.

\section{Stochastic Variational Inference in Topic
  Models \label{sec:topic-models}}

We derived stochastic variational inference, a scalable inference
algorithm that can be applied to a large class of hierarchical
Bayesian models. In this section we show how to use the general
algorithm of \mysec{inference} to derive stochastic variational
inference for two probabilistic topic models: latent Dirichlet
allocation (LDA)~\citep{Blei:2003b} and its Bayesian nonparametric
counterpart, the hierarchical Dirichlet process (HDP) topic
model~\citep{Teh:2006b}.

Topic models are probabilistic models of document collections that use
latent variables to encode recurring patterns of word
use~\citep{Blei:2012}. Topic modeling algorithms are inference
algorithms; they uncover a set of patterns that pervade a collection
and represent each document according to how it exhibits them.  These
patterns tend to be thematically coherent, which is why the models are
called ``topic models.'' Topic models are used for both descriptive
tasks, such as to build thematic navigators of large collections of
documents, and for predictive tasks, such as to aid document
classification.  Topic models have been extended and applied in many
domains.

Topic models assume that the words of each document arise from a
mixture of multinomials. Across a collection, the documents share the
same mixture components (called \textit{topics}). Each document,
however, is associated with its own mixture proportions (called
\textit{topic proportions}). In this way, topic models represent
documents heterogeneously---the documents share the same set of topics,
but each exhibits them to a different degree.  For example, a document
about sports and health will be associated with the sports and health
topics; a document about sports and business will be associated with
the sports and business topics.  They both share the sports topic, but
each combines sports with a different topic.  More generally, this is
called \textit{mixed membership}~\citep{Erosheva:2003}.

The central computational problem in topic modeling is posterior
inference: Given a collection of documents, what are the topics that
it exhibits and how does each document exhibit them?  In practical
applications of topic models, scale is important---these models
promise an unsupervised approach to organizing large collections of
text (and, with simple adaptations, images, sound, and other data).
Thus they are a good testbed for stochastic variational inference.

More broadly, this section illustrates how to use the results from
\mysec{inference} to develop algorithms for specific models.  We will
derive the algorithms in several steps: (1) we specify the model
assumptions; (2) we derive the complete conditional distributions of
the latent variables; (3) we form the mean-field variational family;
(4) we derive the corresponding stochastic inference algorithm.  In
\mysec{experiments}, we will report our empirical study of stochastic
variational inference with these models.

\subsection{Notation}

We follow the notation of \cite{Blei:2003b}.
\begin{itemize}
\item Observations are \textit{words}, organized into documents.  The
  $n$th word in the $d$th document is $w_{dn}$.  Each word is an
  element in a fixed vocabulary of $V$ terms.

\item A \textit{topic} $\beta_k$ is a distribution over the
  vocabulary.  Each topic is a point on the $V-1$-simplex, a positive
  vector of length $V$ that sums to one. We denote the $w$th entry
  in the $k$th topic as $\beta_{kw}$. In LDA there are $K$ topics;
  in the HDP topic model there are an infinite number of topics.

\item Each document in the collection is associated with a vector of
  \textit{topic proportions} $\theta_d$, which is a distribution over
  topics.  In LDA $\theta_d$ is a point on the $K-1$-simplex.  In the
  HDP topic model, $\theta_d$ is a point on the infinite simplex.  (We
  give details about this below in \mysec{hdp}.) We denote the $k$th
  entry of the topic proportion vector $\theta_d$ as $\theta_{dk}$.

\item Each word in each document is assumed to have been drawn from a
  single topic.  The \textit{topic assignment} $z_{dn}$ indexes the
  topic from which $w_{dn}$ is drawn.
\end{itemize}

The only observed variables are the words of the documents.  The
topics, topic proportions, and topic assignments are latent variables.

\subsection{Latent Dirichlet allocation}

\begin{sidewaysfigure}
  \begin{center}
    \includegraphics[scale=0.5]{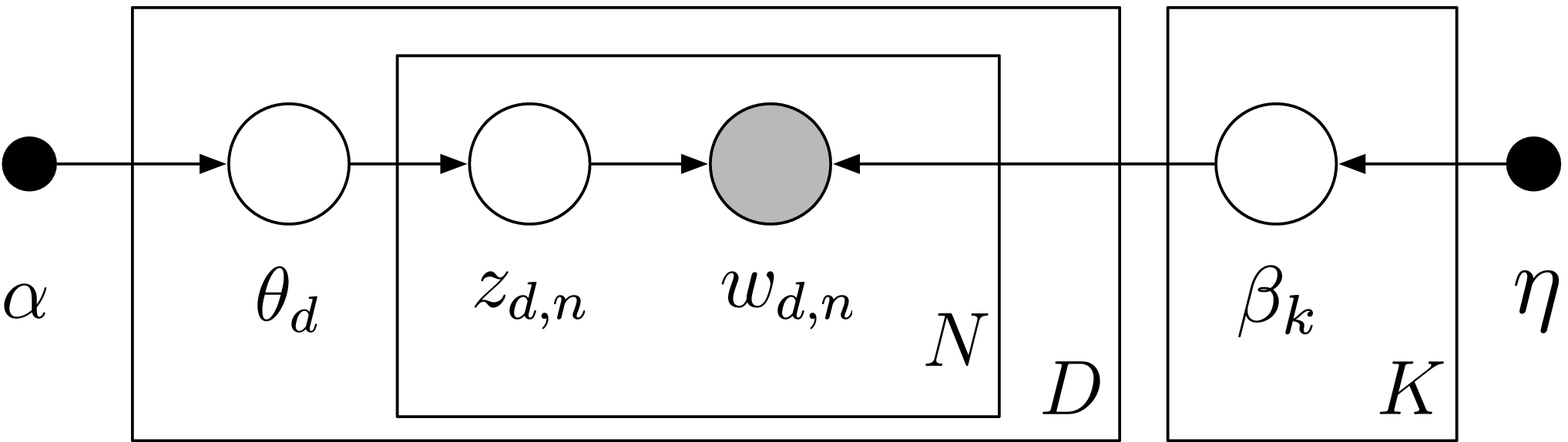}

  \vspace{0.25in}

 \small
  \begin{tabular}{|c|l|l|c|l|}
    \hline
    {Var} & {Type} & {Conditional} &  {Param} & {Relevant Expectations} \\
    \hline \hline
    &&&& \\
    $z_{dn}$ &  Multinomial &
    $\log \theta_{dk} + \log \beta_{k,w_{dn}}$ &
    $\phi_{dn}$ &
    $\E[Z^k_{dn}] = \phi_{dn}^k$ \\
    &&&& \\
    $\theta_d$ & Dirichlet & $\alpha + \textstyle \sum_{n=1}^{N}
    z_{dn}$ & $\gamma_d$ & $\E[\log \theta_{dk}] = \upPsi(\gamma_{dk})
    - \textstyle \sum_{j=1}^{K} \upPsi(\gamma_{dj})$ \\
    &&&&\\
    $\beta_{k}$ & Dirichlet & $\eta + \textstyle \sum_{d=1}^{D}
    \sum_{n=1}^{N} z_{dn}^k w_{dn}$ & $\lambda_k$ & $\E[\log
    \beta_{kv}] = \upPsi(\lambda_{kv}) - \sum_{y=1}^{V} \upPsi(\lambda_{ky})$\\
    &&&&\\
    \hline
  \end{tabular}
  \end{center}

  \caption{(Top) The graphical model representation of Latent
    Dirichlet allocation.  Note that in practice each document $d$ may
    not have the same number of words $N$. (Bottom) In LDA: hidden
    variables, complete conditionals, variational parameters, and
    expected sufficient statistics.  \label{fig:lda-gm}}

\end{sidewaysfigure}

LDA is the simplest topic model.  It assumes that each document
exhibits $K$ topics with different proportions. The generative process
is
\begin{enumerate}
\item Draw topics $\beta_k \sim \dir(\eta,\ldots,\eta)$ 
  for $k \in \{1, \ldots, K\}$.
\item For each document $d \in \{1, \ldots, D\}$:
  \begin{enumerate}
  \item Draw topic proportions $\theta \sim \dir(\alpha,\ldots,\alpha)$.
  \item For each word $w \in \{1, \ldots, N\}$:
    \begin{enumerate}
    \item Draw topic assignment $z_{dn} \sim \mult(\theta_d)$.
    \item Draw word $w_{dn} \sim \mult(\beta_{z_{dn}})$.
    \end{enumerate}
  \end{enumerate}
\end{enumerate}
\myfig{lda-gm} illustrates LDA as a graphical model.

In LDA, each document exhibits the same shared topics but with
different proportions. LDA assumes Dirichlet priors for $\beta_k$ and
$\theta_d$. Dirichlet distributions over the $D$-simplex take $D+1$
parameters, but for simplicity we assume exchangeable Dirichlet
priors; that is, we require that all of these parameters are set to
the same value. (The prior on $\beta_k$ has parameter $\eta$; the
prior on $\theta_d$ has parameter $\alpha$.). We note that
\cite{Blei:2003b} and \cite{Wallach:2009} found improved empirical
performance with non-exchangeable priors.

LDA models an observed collection of documents $\bw = w_{1:D}$, where
each $w_d$ is a collection of words $w_{d,1:N}$.  Analyzing the
documents amounts to posterior inference of $p(\beta, \bm{\theta}, \bz
\g \bw)$.  Conditioned on the documents, the posterior distribution
captures the topics that describe them ($\beta = \beta_{1:K}$), the
degree to which each document exhibits those topics ($\bm{\theta} =
\theta_{1:D}$), and which topics each word was assigned to ($\bz =
z_{1:D,1:N}$).  We can use the posterior to explore large collections
of documents.  \myfig{nyt-topics} illustrates posterior topics found
with stochastic variational inference.

The posterior is intractable to compute~\citep{Blei:2003b}.
Approximating the posterior in LDA is a central computational problem
for topic modeling. Researchers have developed many methods, including
Markov chain Monte Carlo methods~\citep{Griffiths:2004a}, expectation
propagation~\citep{Minka:2002}, and variational
inference~\citep{Blei:2003b,Teh:2006,Asuncion:2009}.  Here we use the
results of \mysec{inference} to develop stochastic inference for LDA.
This scales the original variational algorithm for LDA to massive
collections of documents.\footnote{The algorithm we present was
  originally developed in \citet{Hoffman:2010a}, which is a special
  case of the stochastic variational inference algorithm we developed
  in \mysec{inference}.}

\begin{figure}
  \begin{center}
    \includegraphics[width=\textwidth]{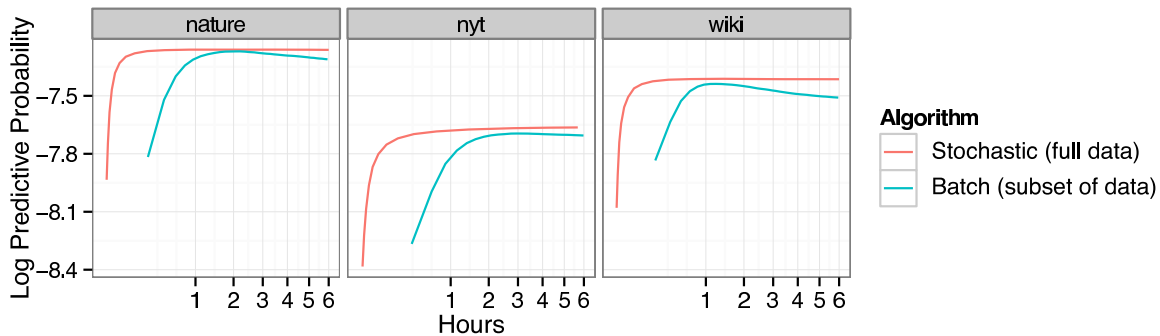}
  \end{center}
  \caption{\label{fig:lda-v-batch} The per-word predictive log
    likelihood for a 100-topic LDA model on three large corpora.
    (Time is on the square root scale.)  Stochastic variational
    inference on the full data converges faster and to a better place
    than batch variational inference on a reasonably sized subset.
    \mysec{experiments} gives the details of our empirical study.}
\end{figure}

\myfig{lda-v-batch} illustrates the performance of 100-topic LDA on
three large collections---\textit{Nature} contains 350K documents,
\textit{New York Times} contains 1.8M documents, and
\textit{Wikipedia} contains 3.8M documents.  (\mysec{experiments}
describes the complete study, including the details of the performance
measure and corpora.)  We compare two inference algorithms for LDA:
stochastic inference on the full collection and batch inference on a
subset of 100,000 documents.  (This is the size of collection that
batch inference can handle.)  We see that stochastic variational
inference converges faster and to a better model. It is both more
efficient and lets us handle the full data set.

\myparagraph{Indicator vectors and Dirichlet distributions.}  Before
deriving the algorithm, we discuss two mathematical details.  These
will be useful both here and in the next section.

First, we represent categorical variables like the topic assignments
$z_{dn}$ and observed words $w_{dn}$ with \textit{indicator
  vectors}.  An indicator vector is a binary vector with a single one.
For example, the topic assignment $z_{dn}$ can take on one of $K$
values (one for each topic).  Thus, it is represented as a $K$-vector
with a one in the component corresponding to the value of the
variable: if $z_{dn}^k = 1$ then the $n$th word in document $d$ is
assigned to the $k$th topic. Likewise, $w_{dn}^v=1$ implies that the
$n$th word in document $d$ is $v$. In a slight abuse of notation, we
will sometimes use $w_{dn}$ and $z_{dn}$ as indices---for example,
if $z_{dn}^k=1$, then $\beta_{z_{dn}}$ refers to the $k$th topic
$\beta_k$.

Second, we review the Dirichlet distribution.  As we described above,
a $K$-dimensional Dirichlet is a distribution on the $K-1$-simplex,
i.e., positive vectors over $K$ elements that sum to one.  It is
parameterized by a positive $K$-vector $\gamma$,
\begin{equation*}
  \dir(\theta; \gamma) =
  \frac{\upGamma\left(\textstyle \sum_{i=1}^{K} \gamma_i\right)}
  {\prod_{i=1}^{K} \upGamma(\gamma_i)}
  \prod_{i=1}^{K} \theta^{\gamma_i-1},
\end{equation*}
where $\upGamma(\cdot)$ is the Gamma function, which is a real-valued
generalization of the factorial function.  The expectation of the
Dirichlet is its normalized parameter,
\begin{equation*}
  % \label{eq:expected-dir}
  \E[\theta_k \g \gamma] = \frac{\gamma_k}{\textstyle \sum_{i=1}^{K}
    \gamma_i}.
\end{equation*}
The expectation of its log uses $\upPsi(\cdot)$, which is the first
derivative of the log Gamma function,
\begin{equation}
  \label{eq:expected-log-dir}
  \E[\log \theta_k \g \gamma] = \upPsi(\gamma_k) - \upPsi\left(\textstyle
    \sum_{i=1}^{K} \gamma_i\right).
\end{equation}
This can be derived by putting the Dirichlet in exponential family
form, noticing that $\log \theta$ is the vector of sufficient
statistics, and computing its expectation by taking the gradient of
the log-normalizer with respect to the natural parameter vector
$\gamma$.

\myparagraph{Complete conditionals and variational distributions.}  We
specify the global and local variables of LDA to place it in the
stochastic variational inference setting of \mysec{inference}.  In
topic modeling, the local context is a document $d$.  The local
observations are its observed words $w_{d,1:N}$. The local hidden
variables are the topic proportions $\theta_d$ and the topic assignments
$z_{d,1:N}$.  The global hidden variables are the topics $\beta_{1:K}$.

Recall from \mysec{inference} that the complete conditional is the
conditional distribution of a variable given all of the other
variables, hidden and observed.  In mean-field variational inference,
the variational distributions of each variable are in the same family
as the complete conditional.

We begin with the topic assignment $z_{dn}$.  The complete
conditional of the topic assignment is a multinomial,
\begin{equation}
  \label{eq:z-cond}
  p(z_{dn} = k \g \theta_d, \beta_{1:K}, w_{dn}) \propto
  \exp\{\log \theta_{dk} + \log \beta_{k, w_{dn}}\}.
\end{equation}
Thus its variational distribution is a multinomial $q(z_{dn}) =
\mult(\phi_{dn})$, where the variational parameter $\phi_{dn}$ is a
point on the $K-1$-simplex.  Per the mean-field approximation, each
observed word is endowed with a different variational distribution for
its topic assignment, allowing different words to be associated with
different topics.

The complete conditional of the topic proportions is a Dirichlet,
\begin{equation}
  \label{eq:theta-cond}
  p(\theta_d \g \bz_d) = \dir\left(\alpha + \textstyle
    \sum_{n=1}^{N} z_{dn}\right).
\end{equation}
Since $z_{dn}$ is an indicator vector, the $k$th element of the
parameter to this Dirichlet is the sum of the hyperparameter $\alpha$
and the number of words assigned to topic $k$ in document $d$.  Note
that, although we have assumed an exchangeable Dirichlet prior, when
we condition on $z$ the conditional $p(\theta_d|z_d)$ is a
non-exchangeable Dirichlet.

With this conditional, the variational distribution of the topic
proportions is also Dirichlet $q(\theta_d) = \dir(\gamma_d)$, where
$\gamma_d$ is a $K$-vector Dirichlet parameter.  There is a different
variational Dirichlet parameter for each document, allowing different
documents to be associated with different topics in different
proportions.

These are local hidden variables.  The complete conditionals only
depend on other variables in the local context (i.e., the document)
and the global variables; they do not depend on variables from other
documents.

Finally, the complete conditional for the topic $\beta_k$ is also a
Dirichlet,
\begin{equation}
  \label{eq:beta-cond}
  p(\beta_k \g \bz, \bw) = \dir\left(\eta + \textstyle \sum_{d=1}^{D}
    \sum_{n=1}^{N}
    z_{dn}^k w_{dn}\right).
\end{equation}
The $v$th element of the parameter to the Dirichlet conditional for
topic $k$ is the sum of the hyperparameter $\eta$ and the number of
times that the term $v$ was assigned to topic $k$.  This is a global
variable---its complete conditional depends on the words and topic
assignments of the entire collection.

The variational distribution for each topic is a $V$-dimensional
Dirichlet,
\begin{equation*}
  q(\beta_k) = \dir(\lambda_k).
\end{equation*}
As we will see in the next section, the traditional variational
inference algorithm for LDA is inefficient with large collections of
documents.  The root of this inefficiency is the update for the topic
parameter $\lambda_k$, which (from \myeq{beta-cond}) requires summing
over variational parameters for every word in the collection.

\myparagraph{Batch variational inference.}

%\begin{algorithm}[tb]
%  \caption{Batch variational Bayes for LDA}
%  \label{alg:batch-lda}
%  \begin{algorithmic}
%    \STATE Initialize $\lambda$ randomly.
%    \WHILE{relative improvement in $\cL(w, \phi, \gamma, \lambda) > 0.00001$}
%    \STATE {\it E step}:
%    \FOR{$d=1$ to $D$}
%    \STATE Initialize $\gamma_{dk}=1$. (The constant 1 is arbitrary.)
%    \REPEAT
%    \STATE Set $\phi_{dwk} \propto
%    \exp\{\Eq[\log \theta_{dk}] + \Eq[\log \beta_{kw}]\}$
%    \STATE Set $\gamma_{dk} = \alpha + \sum_w \phi_{dwk}n_{dw}$
%    \UNTIL{$\frac{1}{K}\sum_k |\mbox{change in}  \gamma_{dk}| < 0.00001$}
%    \ENDFOR
%    \STATE {\it M step}:
%    \STATE Set $\lambda_{kw} = \eta + \sum_d n_{dw} \phi_{dwk}$
%    \ENDWHILE
%  \end{algorithmic}
%\end{algorithm}

With the complete conditionals in hand, we now derive the coordinate
ascent variational inference algorithm, i.e., the batch inference
algorithm of \myfig{classical-vi}.  We form each coordinate update by
taking the expectation of the natural parameter of the complete
conditional.  This is the stepping stone to stochastic variational
inference.

The variational parameters are the global per-topic Dirichlet parameters
$\lambda_{1:K}$, local per-document Dirichlet parameters $\gamma_{1:D}$, and
local per-word multinomial parameters $\phi_{1:D, 1:N}$.  Coordinate ascent
variational inference iterates between updating all of the local
variational parameters (\myeq{coord-local}) and updating the global
variational parameters (\myeq{coord-global}).

We update each document $d$'s local variational in a local coordinate
ascent routine, iterating between updating each word's topic
assignment and the per-document topic proportions,
\begin{eqnarray}
  \phi_{dn}^k &\propto& \exp\left\{\upPsi(\gamma_{dk}) +
  \upPsi(\lambda_{k,w_{dn}}) - \upPsi\left(\textstyle \sum_{v}
  \lambda_{kv}\right)\right\}
  \label{eq:z-update}
  \quad
  \textrm{for } n \in \{1, \ldots, N\} \\
  \gamma_d &=& \alpha + \textstyle \sum_{n=1}^{N} \phi_{dn}.
  \label{eq:theta-update}
\end{eqnarray}
These updates derive from taking the expectations of the natural
parameters of the complete conditionals in \myeq{z-cond} and
\myeq{theta-cond}.  (We then map back to the usual parameterization
of the multinomial.)  For the update on the topic
assignment, we have used the Dirichlet expectations in
\myeq{expected-log-dir}.  For the update on the topic proportions, we
have used that the expectation of an indicator is its probability,
$\E_q[z_{dn}^k] = \phi_{dn}^k$.

After finding variational parameters for each document, we update the
variational Dirichlet for each topic,
\begin{equation}
  \label{eq:lambda-update}
  \lambda_k = \eta + \textstyle \sum_{d=1}^{D} \sum_{n=1}^{N}
  \phi_{dn}^k w_{dn}.
\end{equation}
This update depends on the variational parameters $\phi$ from every document.
%With these local and global updates, \myalg{batch-LDA} presents the
%full algorithm for batch variational inference in LDA.

Batch inference is inefficient for large collections of documents.
Before updating the topics $\lambda_{1:K}$, we compute the local
variational parameters for every document. This is particularly
wasteful in the beginning of the algorithm when, before completing
the first iteration, we must analyze every document with randomly
initialized topics.

\myparagraph{Stochastic variational inference}
\begin{figure}
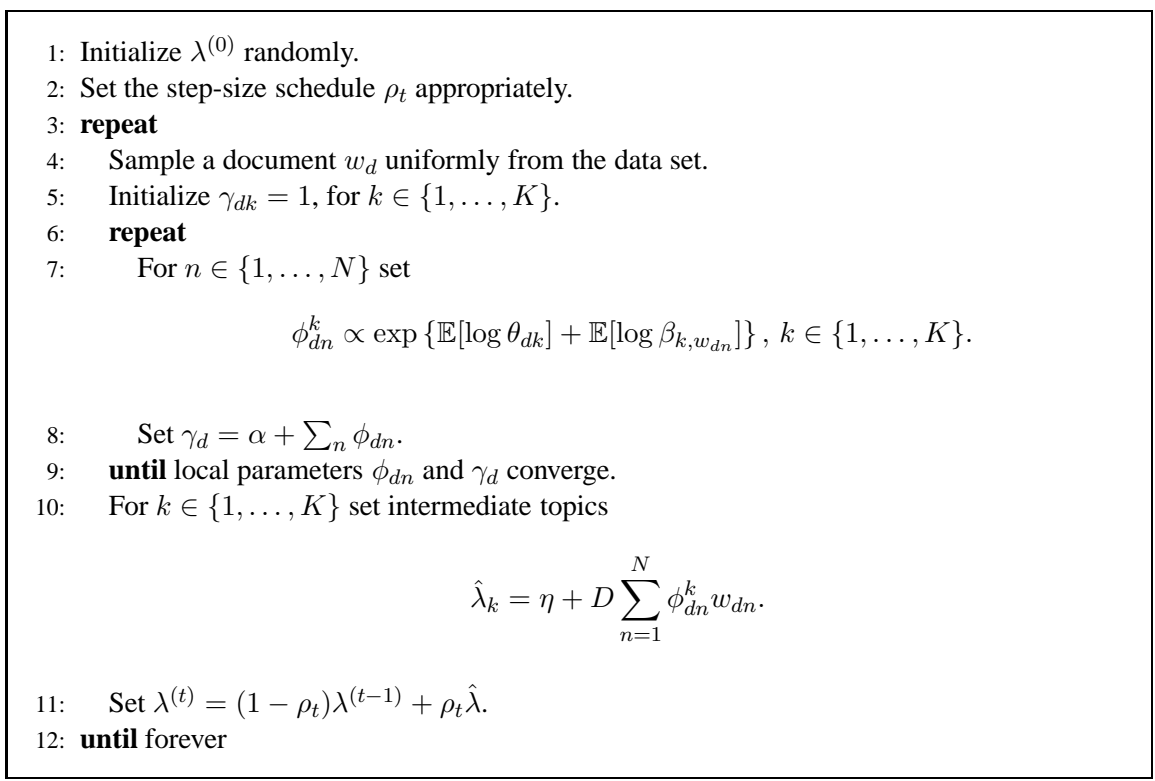

\begin{framed}
   \begin{algorithmic}[1]
     \STATE Initialize $\lambda^{(0)}$ randomly.
     \STATE Set the step-size schedule $\rho_t$ appropriately.
     \REPEAT
     \STATE Sample a document $w_d$ uniformly from the data set.
     \STATE Initialize $\gamma_{dk}=1$, for $k \in \{1, \ldots, K\}$.
     \REPEAT
     \STATE For $n \in \{1, \ldots, N\}$ set
     \[\phi_{dn}^k \propto \exp\left\{\E[\log\theta_{dk}] + \E[\log
       \beta_{k, w_{dn}}]\right\} , \, k \in
     \{1,\ldots,K\}.\]
     \STATE Set $\gamma_{d} = \alpha + \sum_n
     \phi_{dn}$.

     \UNTIL{local parameters $\phi_{dn}$ and $\gamma_d$
       converge}.

     \STATE For $k \in \{1, \ldots, K\}$ set intermediate topics
     \[ \hat{\lambda}_{k} = \eta +
     D\sum_{n=1}^N\phi_{dn}^kw_{dn}. \]
     \STATE Set $\lambda^{(t)} = (1-\rho_t)\lambda^{(t-1)} +
     \rho_t\hat{\lambda}$.
     \UNTIL{forever}
   \end{algorithmic}
\end{framed}
\caption{Stochastic variational inference for LDA.  The relevant
  expectations for each update are found in \myfig{lda-gm}.}
  \label{fig:stoch-lda}
\end{figure}

Stochastic variational inference provides a scalable method for
approximate posterior inference in LDA.  The global variational
parameters are the topic Dirichlet parameters $\lambda_k$; the local
variational parameters are the per-document topic proportion Dirichlet
parameters $\gamma_d$ and the per-word topic assignment multinomial
parameters $\phi_{dn}$.

We follow the general algorithm of \myfig{stoch-vi}.  Let
$\lambda^{(t)}$ be the topics at iteration $t$.  At each iteration we
sample a document $d$ from the collection.  In the local phase, we
compute optimal variational parameters by iterating between updating
the per-document parameters $\gamma_d$ (\myeq{theta-update}) and
$\phi_{d,1:N}$ (\myeq{z-update}).  This is the same subroutine as in
batch inference, though here we only analyze one randomly chosen document.

In the global phase we use these fitted local variational parameters
to form intermediate topics,
\begin{equation*}
  \hat{\lambda}_k = \eta + D \textstyle \sum_{n=1}^{N} \phi_{dn}^k
  w_{dn}.
\end{equation*}
This comes from applying \myeq{lambda-update} to a hypothetical corpus
containing $D$ replicates of document $d$.  We then set the topics at
the next iteration to be a weighted combination of the intermediate
topics and current topics,
\begin{equation*}
  \lambda_k^{(t+1)} = (1-\rho_t) \lambda_k^{(t)} + \rho_t\hat{\lambda}_k.
\end{equation*}
\myfig{stoch-lda} gives the algorithm for stochastic variational
inference for LDA.\footnote{This algorithm, as well as the algorithm
  for the HDP, specifies that we initialize the topics $\lambda_k$
  randomly.  There are many ways to initialize the topics.  We use an
  exponential distribution,
  \begin{equation*}
    \lambda_{kv} - \eta \sim \textrm{Exponential}(D * 100 / (KV)).
  \end{equation*}
  This gives a setting of $\lambda$ similar to the one we would get by
  applying \myeq{lambda-update} after randomly assigning words to
  topics in a corpus of size $D$ with 100 words per document.}

% !!! we should say s/t about convergence somewhere (not needed for
% submission)  i think this should go in the previous section.

% !!! these plots should change to # documents at log scale on the x
% axis (not needed for submission)

\subsection{Bayesian nonparametric topic models with the HDP}
\label{sec:hdp}

Stochastic inference for LDA lets us analyze large collections of
documents.  One limitation of LDA, however, is that the number of
topics is fixed in advance.  Typically, researchers find the ``best''
number of topics with cross-validation~\citep{Blei:2003b}.  However,
for very large data this approach is not practical.  We can address
this issue with a Bayesian nonparametric topic model, a model where
the documents themselves determine the number of topics.

We derive stochastic variational inference for the Bayesian
nonparametric variant of LDA, the hierarchical Dirichlet process (HDP)
topic model.  Like LDA, the HDP topic model is a mixed-membership
model of text collections.  However, the HDP assumes an ``infinite''
number of topics.  Given a collection of documents, the posterior
distribution of the hidden structure determines how many topics are
needed to describe them.  Further, the HDP is flexible in that it
allows future data to exhibit new and previously unseen topics.

More broadly, stochastic variational inference for the HDP topic model
demonstrates the possibilities of stochastic inference in the context
of Bayesian nonparametric statistics.  Bayesian nonparametrics gives
us a collection of flexible models---mixture models, mixed-membership
models, factor models, and models with more complex structure---which
grow and expand with data~\citep{BNP:2010}.  Flexible and expanding
models are particularly important when analyzing large data sets,
where it is prohibitive to search for a specific latent structure
(such as a number of topics or a tree structure of components) with
cross-validation.  Here we demonstrate how to use stochastic inference
in the context of a simple Bayesian nonparametric topic model.  In
other work, we built on this algorithm to give scalable inference
methods for Bayesian nonparametric models of topic
correlation~\citep{Paisley:2012b} and tree structures of
topics~\citep{Paisley:2012c}.

This section is organized as follows.  We first give some background
on the Dirichlet process and its definition via Sethuraman's stick
breaking construction, which is a distribution on the infinite
simplex.  We then show how to use this construction to form the HDP
topic model and how to use stochastic variational inference to
approximate the posterior.\footnote{This algorithm first appeared
  in~\cite{Wang:2011}.  Here we place it in the more general context
  of \mysec{inference} and relate it to stochastic inference for LDA.}

% dmb: i wrote and removed this paragraph---keep it positive :)

% LDA requires setting the number of topics.  In many applications,
% however, we would like to determine that number from the data.  For
% example, we can use cross validation and a measure of model fitness to
% find a good setting.  This can be expensive in large data sets---each
% fold requires fitting a new LDA model for multiple numbers of
% topics. And cross-validation cannot be considered if we want to fit a
% model to streaming data, which online inference for LDA allows us to
% do.  Finally, LDA assumes that all future data will exhibit the same
% number and set of topics that were fixed from the start.

\myparagraph{The stick-breaking construction of the Dirichlet
  process.}  Bayesian nonparametric (BNP) methods use distributions of
distributions, placing flexible priors on the shape of the
data-generating density function.  BNP models draw a distribution from
that prior and then independently draw data from that random
distribution.  Data analysis proceeds by evaluating the posterior
distribution of the (random) distribution from which the data were
drawn.  Because of the flexible prior, that posterior can potentially
have mass on a wide variety of distribution shapes.  For a reviews of
BNP methods, see the edited volume of~\cite{BNP:2010} and the tutorial
of~\cite{Gershman:2012}.

% dmb: john, don't change "need not be discrete" back to "which may be
% discrete".  i also avoid "need not"---it sucks in most writing
% circumstances.  but i think it's the right move here.

The most common BNP prior is the \textit{Dirichlet process} (DP).  The
Dirichlet process is parameterized by a \textit{base distribution}
$G_0$ (which may be either continuous or discrete) and a non-negative scaling
factor $\alpha$.  These are used to form a distribution over discrete
distributions, i.e., over distributions that place their mass on a
countably infinite set of atoms.  The locations of the atoms are
independently drawn from the base distribution $G_0$ and the closeness
of the probabilities to $G_0$ is determined by the scaling factor
$\alpha$. When $\alpha$ is small, more mass is placed on fewer atoms,
and the draw will likely look very different from $G_0$; when $\alpha$
is large, the mass is spread around many atoms, and the draw will more
closely resemble the base distribution.

There are several representations of the Dirichlet process. For
example, it is a normalized gamma process~\citep{Ferguson:1973}, and
its marginalization gives the Chinese restaurant
process~\citep{Pitman:2002}. We will focus on its definition via
Sethuraman's stick breaking construction~\citep{Sethuraman:1994}.  The
stick-breaking construction explicitly defines the distribution of the
probabilities that make up a random discrete distribution.  It is the
gateway to variational inference in Bayesian nonparametric
models~\citep{Blei:2006e}.

Let $G \sim \textrm{DP}(\alpha, G_0)$ be drawn from a Dirichlet
process prior. It is a discrete distribution with mass on an infinite
set of atoms.  Let $\beta_k$ be the atoms in this distribution and
$\sigma_k$ be their corresponding probabilities.  We can write $G$ as
\begin{equation*}
  G = \sum_{k=1}^{\infty} \sigma_k \delta_{\beta_k}.
\end{equation*}
The atoms are drawn independently from $G_0$.  The stick-breaking
construction specifies the distribution of their probabilities.

The stick-breaking construction uses an infinite collection of
beta-distributed random variables. Recall that the beta is a
distribution on $(0,1)$ and define the following collection,
\begin{equation*}
  v_i \sim \Bet(1, \alpha) \quad i \in \{1, 2, 3, \ldots\}.
\end{equation*}
These variables combine to form a point on the infinite simplex.
Imagine a stick of unit length.  Break off the proportion of the stick
given by $v_1$, call it $\sigma_1$, and set it aside.  From the
remainder (of length $1 - \sigma_1$) break off the proportion given by
$v_2$, call it $\sigma_2$, and set it aside. The remainder of the
stick is now $1-\sigma_2-\sigma_1 = (1-v_1)(1-v_2)$.  Repeat this
process for the infinite set of $v_i$.  The resulting stick lengths
$\sigma_i$ will sum to one.

More formally, we define the function $\sigma_i$ to take the
collection of realized $v_i$ variables and to return the stick length
of the $i$th component,
\begin{equation*}
  \sigma_i(\bv) = v_i \textstyle \prod_{j=1}^{i-1} (1 - v_j),
\end{equation*}
and note that $\sum_{i=1}^{\infty} \sigma_i(\bv) = 1$.  We call $v_i$
the $i$th \textit{breaking proportion}.

Combining these steps, we form the distribution $G$ according to the
following process,
\begin{eqnarray*}
  \beta_i &\sim& G_0 \quad i \in \{1,2,3, \ldots\} \\
  v_i &\sim& \Bet(1,\alpha) \quad i \in \{1, 2, 3, \ldots\} \\
  G &=& \textstyle \sum_{i=1}^{\infty} \sigma_i(\bv)
  \delta_{\beta_i}.
\end{eqnarray*}
In the random distribution $G$ the $i$th atom $\beta_i$ is an
independent draw from $G_0$ and it has probability given by the $i$th
stick length $\sigma_i(\bv)$. \cite{Sethuraman:1994} showed that the
distribution of $G$ is $\textrm{DP}(\alpha, G_0)$.

The most important property of $G$ is the ``clustering'' property.
Even though $G$ places mass on a countably infinite set of atoms, $N$
draws from $G$ will tend to exhibit only a small number of them.  (How
many depends on the scalar $\alpha$, as we described above.)
Formally, this is most easily seen via other perspectives on the
DP~\citep{Ferguson:1973,Blackwell:1973,Pitman:2002}, though it can be
seen intuitively with the stick-breaking construction.  The intuition
is that as $\alpha$ gets smaller more of the stick is absorbed in the
first break locations because the breaking proportions are drawn from
$\textrm{Beta}(1, \alpha)$.  Thus, those atoms associated with the
first breaks of the stick will have larger mass in the distribution
$G$, and that in turn encourages draws from the distribution to
realize fewer individual atoms.  In general, the first break locations
tend to be larger than the later break locations.  This property is
called \textit{size biasedness}.

% dmb: commented out below, per john's suggestion that we need to get
% to the point.  (and i agree.)

% The clustering property motivates the Dirichlet process mixture model,
% where the atoms represent mixture components and the data are drawn
% using components drawn from $G$~\citep{Antoniak:1974,Rasmussen:2002}.
% Given data, the posterior distribution of a DP mixture gives a form of
% model selection: the data are attributed to a small number of
% components, where the number of components is not known in advance.
% Further, DP mixtures are more flexible than most types of model
% selection because the predictive distribution allows new data to be
% explained with previously unseen components.

\myparagraph{The HDP topic model.}  We now construct a Bayesian
nonparametric topic model that has an ``infinite'' number of topics.
The hierarchical Dirichlet process topic model couples
%% a two-level Dirichlet process
a set of document-level DPs via a single top-level
DP~\citep{Teh:2006b}.  The base distribution $H$ of the top-level DP
is a symmetric Dirichlet over the vocabulary simplex---its atoms are
topics.  We draw once from this DP, $G_0 \sim \textrm{DP}(\omega, H)$.
In the second level, we use $G_0$ as a base measure to a
document-level DP, $G_d \sim \textrm{DP}(\alpha, G_0)$.  We draw the
words of each document $d$ from topics from $G_d$.  The consequence of
this two-level construction is that all documents share the same
collection of topics but exhibit them with different proportions.

\begin{sidewaysfigure}
\begin{center}
  \includegraphics[scale=0.5]{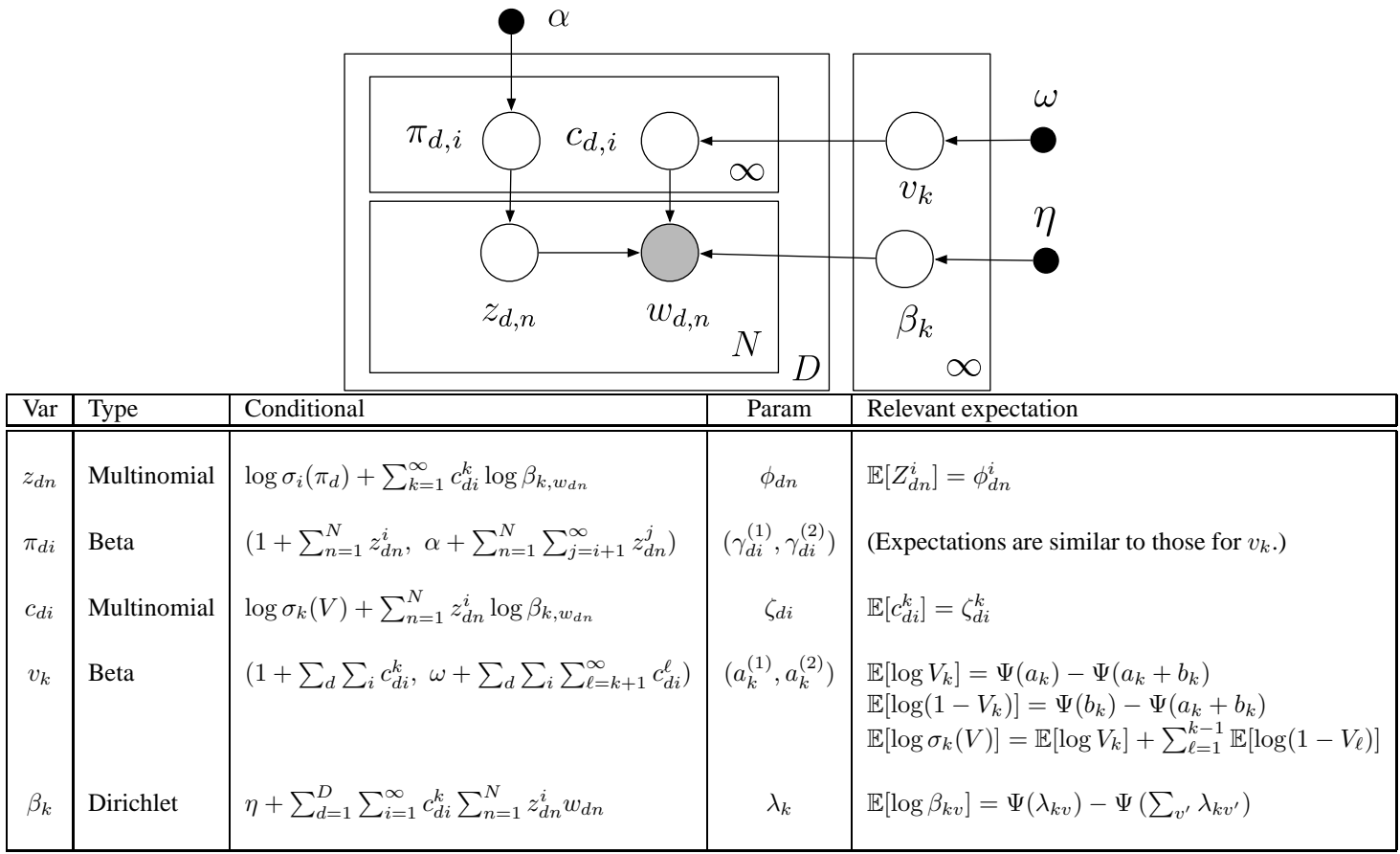}
  \small
  \begin{tabular}{|c|l|l|c|l|}
    \hline
    {Var} & {Type} & {Conditional} &  {Param} & {Relevant expectation} \\
    \hline \hline
    &&&& \\
    $z_{dn}$ &
    Multinomial &
    $\log \sigma_i(\bm{\pi}_d) + \sum_{k=1}^\infty c_{di}^k \log \beta_{k, w_{dn}}$ &
    $\phi_{dn}$ &
    $\E[Z_{dn}^i] = \phi_{dn}^i$ \\

    &&&&\\

    $\pi_{di}$&
    Beta &
    $(1 + \sum_{n=1}^{N} z_{dn}^i, \,\, \alpha +
    \sum_{n=1}^{N} \sum_{j=i+1}^{\infty} z_{dn}^j)$ &
    $(\gamma_{di}^{(1)}, \gamma_{di}^{(2)})$ &
    (Expectations are similar to those for $v_k$.) \\

    &&&&\\

    $c_{di}$ &
    Multinomial &
    $\log \sigma_k(\vct{V}) + \sum_{n=1}^{N} z_{dn}^i \log
    \beta_{k,w_{dn}}$ &
    $\zeta_{di}$ &
    $\E[c_{di}^k] = \zeta_{di}^k$ \\

    &&&&\\

    $v_k$ & Beta & $( 1 + \sum_{d} \sum_{i} c_{di}^k, \, \, \omega +
    \sum_{d} \sum_{i} \sum_{\ell=k+1}^\infty c_{di}^\ell )$ &
    $( a_k^{(1)}, a_k^{(2)} )$ & $\E[\log V_k] = \upPsi(a_k)
    - \upPsi(a_k +
    b_k)$ \\
    &&&& $\E[\log (1-V_k)] = \upPsi(b_k) - \upPsi(a_k + b_k)$ \\
    &&&& $\E[\log \sigma_k(\vct{V})] = \E[\log V_k] +
    \sum_{\ell=1}^{k-1} \E[\log(1 - V_\ell)]$ \\

    &&&&\\

    $\beta_{k}$ &
    Dirichlet &
    $\eta + \textstyle \sum_{d=1}^{D} \sum_{i=1}^{\infty} c_{di}^k \sum_{n=1}^{N}
    z_{dn}^i w_{dn}$ &
    $\lambda_k$ &
    $\E[\log \beta_{kv}] = \upPsi(\lambda_{kv})
    - \upPsi\left(\textstyle \sum_{v'} \lambda_{kv'}\right)$ \\

    &&&&\\

    \hline
  \end{tabular}
\end{center}
\caption{\label{fig:hdp-gm} A graphical model for the HDP topic model, and
  a summary of its variational inference algorithm.}
\end{sidewaysfigure}

We construct the HDP topic model using a stick-breaking construction
at each level---one at the document level and one at the corpus
level.\footnote{ See the original HDP paper of~\cite{Teh:2006b} for
  other constructions of the HDP---the random measure construction,
  the construction by the Chinese restaurant franchise, and an
  alternative stick-breaking construction. This construction was
  mentioned by \cite{Fox:2008}.  We used it for the HDP in
  \cite{Wang:2011}.}  The generative process of the HDP topic model is
as follows.
\begin{enumerate}
\item Draw an infinite number of topics, $\beta_k \sim \dir(\eta)$
  for $k \in \{1, 2, 3, \ldots\}$.
\item Draw corpus breaking proportions, $v_k \sim \textrm{Beta}(1,
  \omega)$ for $k \in \{1, 2, 3, \ldots\}$.
\item For each document $d$:
  \begin{enumerate}
  \item Draw document-level topic indices, $c_{di} \sim \mult(\sigma(\bv))$ for
    $i \in \{1, 2, 3, \ldots\}$.
  \item Draw document breaking proportions, $\pi_{di} \sim \textrm{Beta}(1,
   \alpha)$ for $i \in \{1, 2, 3, \ldots \}$.
  \item For each word $n$:
    \begin{enumerate}
    \item Draw topic assignment $z_{dn} \sim \mult(\sigma(\bm{\pi}_d))$.
    \item Draw word $w_n \sim \mult(\beta_{c_{d, z_{dn}}})$.
    \end{enumerate}
  \end{enumerate}
\end{enumerate}
\myfig{hdp-gm} illustrates this process as a graphical model.

% Recall that the stick lengths are size-biased, the first breaks of
% the stick will tend to be larger than subsequent breaks.  We do not
% want to assume that what tends to be the main topic in one document
% (i.e., biggest stick length at the document level) is the same as
% the main topic from another. So, for each cell of the topic
% proportions we randomly draw a \textit{topic pointer}, which indexes
% the topic that cell is associated with.  This is where the corpus
% stick comes in.
% 
% % (In LDA, the topic pointers mapped directly to the topics, i.e.,
% % $\theta_k$ refers to the $k$th topic.)
% 
% There are an infinite number of topics, and so an infinite number of
% possible values for the topic pointers.  The distribution over topic
% pointers comes from the corpus stick, which captures the relative
% frequencies of topics in documents.  If one topic appears in many
% documents then it will have a high probability in the corpus stick.
% Finally, as for LDA, the topics themselves---one for each cell in
% the corpus stick---are drawn from an exchangeable Dirichlet.
% 
% \myfig{hdp-gm} illustrates a schematic if the two levels of stick
% breaking constructions.  See the original HDP paper
% of~\cite{Teh:2006b} for other constructions of the HDP---the random
% measure construction, the construction by the Chinese restaurant
% franchise, and an alternative stick-breaking construction.  This
% stick-breaking construction, which facilitates variational
% inference, was introduced in~\cite{Fox:2008}.

In this construction, topics $\beta_k$ are drawn as in LDA (Step 1).
Corpus-level breaking proportions $\bm{v}$ (Step 2) define a
probability distribution on these topics, which indicates their
relative prevalence in the corpus.  At the document level, breaking
proportions $\pi_{d}$ create a set of probabilities (Step 3b) and
topic indices $c_{d}$, drawn from $\sigma(\bm{v})$, attach each
document-level stick length to a topic (Step 3a).  This creates a
document-level distribution over topics, and words are then drawn as
for LDA (Step 3c).

% Together, these define $G_0 = \sum_{k=1}^{\infty} \sigma_k(\bm{v})
% \delta_{\beta_k}$, which is the draw from the top-level DP.

% Calling this top Dirichlet process $G_0$, from Sethuraman we can draw
% each document's distribution in exactly the same way. However, since
% the base $G_0$ has been discretized, we can use a hidden indicator
% variable $c_{di}$ to point to the atom indexed by $G_0$ associated
% with the $i$th stick break for the $d$th document (Step 3a). The
% breaking proportions in Step 3b define these stick lengths through the
% function $\sigma(\bpi_d)$. Words are then drawn as in LDA.

% Given a document collection, the posterior distribution of the latent
% variables gives estimates of a set of topics and how the documents
% exhibit them.  Unlike LDA, the number of topics is determined by the
% data.  And, in the predictive distribution, a new document can exhibit
% new topics.

The posterior distribution of the HDP topic model gives a
mixed-membership decomposition of a corpus where the number of topics
is unknown in advance and unbounded.  However, it is not possible to
compute the posterior.  Approximate posterior inference for BNP models
in general is an active field of
research~\citep{Escobar:1995,Neal:2000,Blei:2006e,Teh:2007b}.

The advantage of our construction over others is that it meets the
conditions of \mysec{inference}.  All the complete conditionals are in
exponential families in closed form, and it neatly separates global
variables from local variables.  The global variables are topics and
corpus-level breaking proportions; the local variables are
document-level topic indices and breaking proportions.  Following the
same procedure as for LDA, we now derive stochastic variational
inference for the HDP topic model.

% Here we derive stochastic variational inference for the HDP topic
% model. Stochastic variational inference gives a scalable inference
% algorithm for fitting HDP topic models. We first compute the set of
% conditional distributions, which will allow us to define the
% appropriate variational distributions and their respective updates
% for inference.

\myparagraph{Complete conditionals and variational distributions.}
We form the complete conditional distributions of all variables in the
HDP topic model. We begin with the latent indicator variables,
\begin{eqnarray}
p(z_{dn}^i=1|\bpi_d,\beta_{1:K},w_{dn},{c}_d) 
&\propto& 
\exp\{\log \sigma_i(\bpi_d) + \textstyle\sum_{k=1}^{\infty}c_{di}^k\log \beta_{k,w_{dn}}  \}, \\
p(c_{di}^k=1|\bv,\beta_{1:K},\bw_d,\bz_d) &\propto& \exp\{\log \sigma_k(\bv) + \textstyle\sum_{n=1}^N z_{dn}^i \log \beta_{k,w_{dn}}\}.
\end{eqnarray}
Note the interaction between the two levels of latent indicators.  In
LDA the $i$th component of the topic proportions points to the $i$th
topic.  Here we must account for the topic index $c_{di}$, which is a
random variable that points to one of the topics.

This interaction between indicators is also seen in the conditionals
for the topics,
\begin{equation*}
  p(\beta_k|\bz,{c},\bw) = \dir\left(\eta +
    \textstyle\sum_{d=1}^D\sum_{i=1}^{\infty}c_{di}^k\sum_{n=1}^N
    z_{dn}^i w_{dn}\right).
\end{equation*}
The innermost sum collects the sufficient statistics for words in the
$d$th document that are allocated to the $i$th local component
index. However, these statistics are only kept when the $i$th topic
index $c_{di}$ points to the $k$th global topic.

The full conditionals for the breaking proportions follow those of a
standard stick-breaking construction \citep{Blei:2006e},
\begin{eqnarray*}
  p(v_k|{c}) &=&
  \Bet\left(1+\textstyle\sum_{d=1}^D\sum_{i=1}^{\infty} c_{di}^k\,
    ,\,\omega + \textstyle\sum_{d=1}^D\sum_{i=1}^{\infty}\sum_{j>k}
    c_{di}^j \right),\\
  p(\pi_{di}|z_d) &=& \Bet\left(1+\textstyle\sum_{n=1}^N z_{dn}^i\,
    ,\, \alpha + \textstyle\sum_{n=1}^N \sum_{j>i} z_{dn}^j  \right).
\end{eqnarray*}

The complete conditionals for all the latent variables are all in the
same family as their corresponding distributions in the generative
process.  Accordingly, we will define the variational distributions to
be in the same family.  However, the main difference between BNP
models and parametric models is that BNP models contain an infinite
number of hidden variables.  These cannot be completely represented in
the variational distribution as this would require optimizing an
infinite number of variational parameters.  We solve this problem by
truncating the variational distribution~\citep{Blei:2006e}.  At the
corpus level, we truncate at $K$, fitting posteriors to $K$ breaking
points, $K$ topics, and allowing the topic pointer variables to take
on one of $K$ values. At the document level we truncate at $T$,
fitting $T$ breaking proportions, $T$ topic pointers, and letting the
topic assignment variable take on one of $T$ values.  Thus the
variational family is,
\begin{equation*}
  q(\beta, v, \bm{z}, \bm{\pi}) =
  \left(\prod_{k=1}^{K} q(\beta_k \g \lambda_k) q(v_k \g a_k) \right)
  \left(\prod_{d=1}^{D} \prod_{i=1}^{T} q(c_{di} \g
    \zeta_{di}) q(\pi_{di} \g \gamma_{di}) \prod_{n=1}^{N}
    q(z_{dn} \g \phi_{dn})\right)
\end{equation*}

We emphasize that this is not a finite model.  With truncation levels
set high enough, the variational posterior will use as many topics as
the posterior needs, but will not necessarily use all $K$ topics to
explain the observations.  (If $K$ is set too small then the truncated
variational distribution will use all of the topics, but this problem
can be easily diagnosed and corrected.)  Further, a particular
advantage of this two-level stick-breaking distribution is that the
document truncation $T$ can be much smaller than $K$.  Though there
may be hundreds of topics in a large corpus, we expect each document
will only exhibit a small subset of them.

\myparagraph{Stochastic variational inference for HDP topic models.}
From the complete conditionals, batch variational inference proceeds
by updating each variational parameter using the expectation of its
conditional distribution's natural parameter.  In stochastic
inference, we sample a data point, update its local parameters as for
batch inference, and then update the global variables.

To update the global topic parameters, we again form intermediate
topics with the sampled document's optimized local parameters,
\begin{equation*}
  \hat{\lambda}_k = \eta + \textstyle
  D\sum_{i=1}^T\E_q[c_{di}^k]\sum_{n=1}^N \E_q[z_{dn}^i] w_{dn}.
\end{equation*}
We then update the global variational parameters by taking a step in
the direction of the stochastic natural gradient
\begin{equation*}
 \lambda^{(t+1)} = (1-\rho_t)\lambda^{(t)} + \rho_t\hat{\lambda}_k.
\end{equation*}
This mirrors the update for LDA.

\begin{figure}
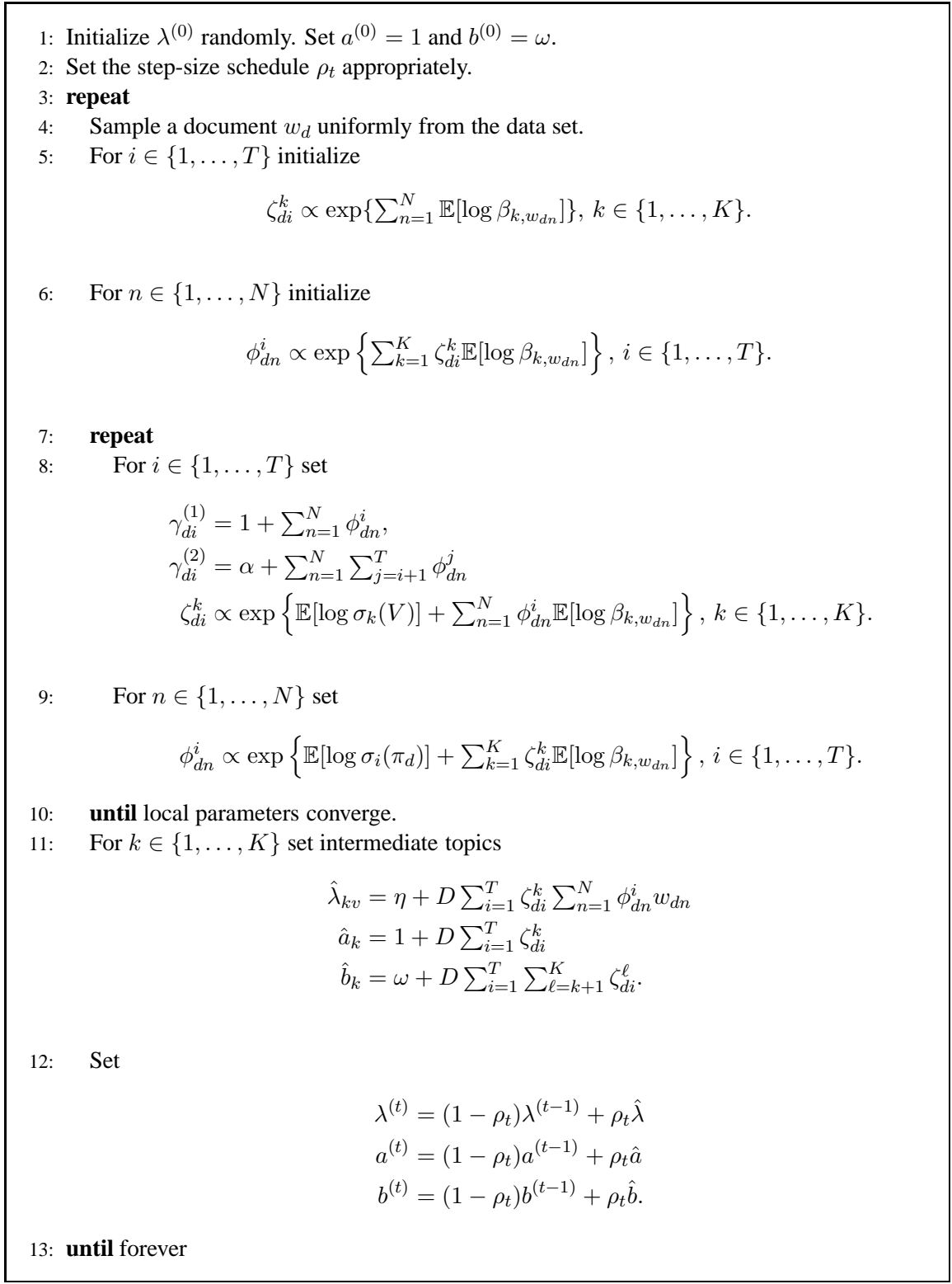

\begin{framed}
   \begin{algorithmic}[1]
     \STATE Initialize $\lambda^{(0)}$ randomly. Set $a^{(0)}=1$ and
     $b^{(0)}=\omega$.

     \STATE Set the step-size schedule $\rho_t$ appropriately.

     \REPEAT

     \STATE Sample a document $w_d$ uniformly from the data set.

     \STATE For $i \in \{1, \ldots, T\}$ initialize
     \[
       \zeta_{di}^k \textstyle \propto \exp\{\sum_{n=1}^{N}
       \E[\log \beta_{k,w_{dn}}]\},\,k \in \{1,\ldots, K\}.
       \]
     \STATE For $n \in \{1,\ldots,N\}$ initialize
     \[\phi_{dn}^i \propto \textstyle \exp\left\{\sum_{k=1}^K \zeta_{di}^k
       \E[\log \beta_{k,w_{dn}}]\right\},\, i \in \{1,\ldots,T\}.\]
     \REPEAT
     \STATE For $i \in \{1,\ldots, T\}$ set
     \begin{align*}
       \gamma_{di}^{(1)} &= \textstyle 1 + \sum_{n=1}^N \phi_{dn}^i,\\
      \gamma_{di}^{(2)} &= \textstyle \alpha + \sum_{n=1}^N \sum_{j=i+1}^T
      \phi_{dn}^j \\
       \zeta_{di}^k & \textstyle \propto \exp\left\{ \E[\log
         \sigma_k(\vct{V})] +
       \sum_{n=1}^{N} \phi_{dn}^i \E[\log \beta_{k,w_{dn}}]\right\}, \, k
       \in \{1,\ldots, K\}.
     \end{align*}
     \STATE For $n \in \{1,\ldots,N\}$ set
     \[\textstyle \phi_{dn}^i \propto \exp\left\{\E[\log \sigma_i
       (\bm{\pi}_d)] +
     \sum_{k=1}^K \zeta_{di}^k \E[\log \beta_{k,w_{dn}}]\right\}, \,
   i \in \{1,\ldots,T\}.\]
     \UNTIL{local parameters converge.}
     \STATE For $k \in \{1, \ldots,K\}$ set intermediate topics
     \begin{align*}
       \hat{\lambda}_{kv} & = \textstyle \eta + D
       \sum_{i=1}^T\zeta_{di}^k\sum_{n=1}^N
       \phi_{dn}^i w_{dn} \\
       \hat{a}_k & =  \textstyle 1 + D \sum_{i=1}^T \zeta_{di}^k \\
       \hat{b}_k & = \textstyle \omega + D  \sum_{i=1}^T \sum_{\ell=k+1}^K
       \zeta_{di}^\ell.
   \end{align*}
   \STATE Set
   \begin{align*}
     \lambda^{(t)} &= (1-\rho_t)\lambda^{(t-1)} + \rho_t\hat{\lambda} \\
     a^{(t)} &= (1-\rho_t)a^{(t-1)} + \rho_t\hat{a} \\
     b^{(t)} &= (1-\rho_t)b^{(t-1)} + \rho_t\hat{b}.
   \end{align*}
     \UNTIL{forever}
   \end{algorithmic}
\end{framed}
\caption{Stochastic variational inference for the HDP topic model.
  The corpus-level truncation is $K$; the document-level truncation as
  $T$. Relevant expectations are found in \myfig{hdp-gm}.}
  \label{fig:stoch-hdp}
\end{figure}

The other global variables in the HDP are the corpus-level breaking
proportions $v_k$, each of which is associated with a set of beta
parameters $a_k = ( a_k^{(1)}, a_k^{(2)} )$ for its
variational distribution.  Using the same randomly selected document
and optimized variational parameters as above, first construct the
two-dimensional vector
\begin{equation*}
  \hat{a}_k = \left( 1+\textstyle D\sum_{i=1}^T
    \E_q[c_{di}^k]\, ,\,\omega + \textstyle
    D\sum_{i=1}^T\sum_{j=k+1}^K \E_q[c_{di}^j]\right).
\end{equation*}
Then, update the parameters
\begin{equation*}
 a_k^{(t+1)} =  (1-\rho_t)a_k^{(t)} + \rho_t \hat{a}_k.
\end{equation*}
Note that we use the truncations $K$ and $T$.  \myfig{hdp-gm}
summarizes the complete conditionals, variational parameters, and
relevant expectations for the full algorithm.  \myfig{stoch-hdp} gives
the stochastic variational inference algorithm for the HDP topic
model.

\begin{figure}
  \begin{center}
    \includegraphics[width=\textwidth]{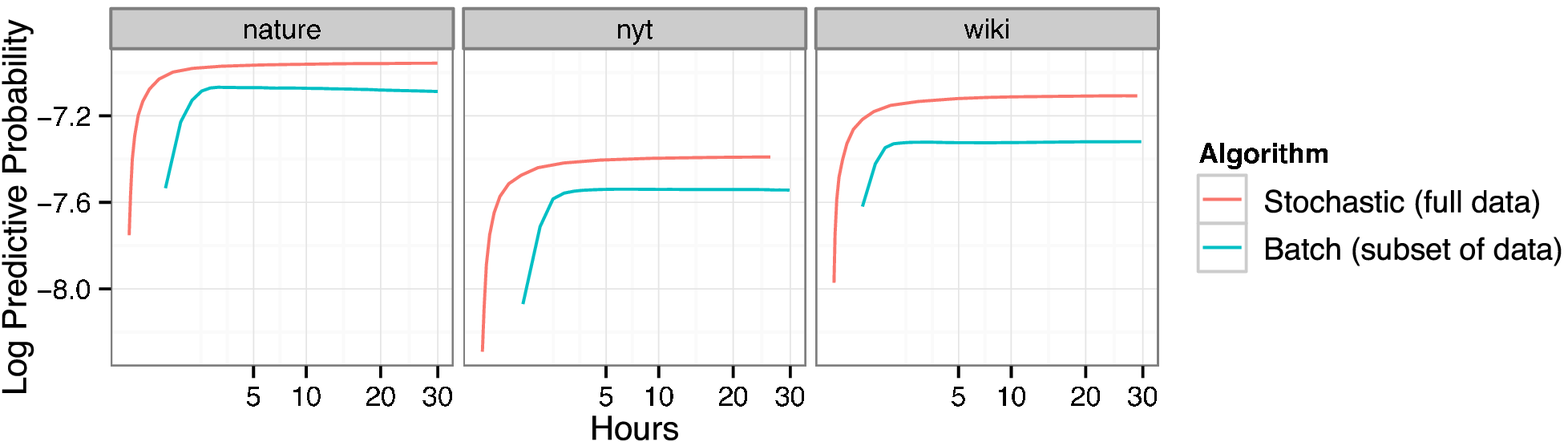}
  \end{center}
  \caption{\label{fig:hdp-v-batch} The per-word predictive log
    likelihood for an HDP model on three large corpora.  (Time is on
    the square root scale.)  As for LDA, stochastic variational
    inference on the full data converges faster and to a better place
    than batch variational inference on a reasonably sized subset.
    \mysec{experiments} gives the details of our empirical study.}
\end{figure}

\myparagraph{Stochastic inference versus batch inference for the HDP.}
\myfig{hdp-v-batch} illustrates the performance of the HDP topic model
on the same three large collections as in \myfig{lda-v-batch}.  As with
LDA, stochastic variational inference for the HDP converges faster and
to a better model.

\section{Empirical Study \label{sec:experiments}}

\newcommand{\test}[1]{\mbox{$#1$}^{\small \mbox{test}}}

In this section we study the empirical performance and effectiveness
of stochastic variational inference for latent Dirichlet allocation
(LDA) and the hierarchical Dirichlet process (HDP) topic model.  With
these algorithms, we can apply and compare these models with very
large collections of documents.  We also investigate how the
forgetting rate $\kappa$ and mini-batch size $S$ influence the
algorithms.  Finally, we compare stochastic variational inference to
the traditional batch variational inference algorithm.\footnote{We
  implemented all algorithms in Python using the NumPy and SciPy
  packages, making the implementations as similar as possible.  Links
  to these implementations are available on the web at
  \\\url{http://www.cs.princeton.edu/~blei/topicmodeling.html}.}

\myparagraph{Data.}  We evaluated our algorithms on three collections of
documents.  For each collection, we computed a vocabulary by removing
stop words, rare words, and very frequent words.  The data are as
follows.
\begin{itemize}
\item {\it Nature}: This collection contains 350,000 documents from
  the journal {\it Nature} (spanning the years 1869--2008). After
  processing, it contains 58M observed words from a vocabulary of 4,200
  terms.
\item {\it New York Times}: This collection contains 1.8M documents
  from the \textit{New York Times} (spanning the years
  1987--2007). After processing, this data contains 461M observed
  words from a vocabulary of 8,000 terms.
\item {\it Wikipedia}: This collections contains 3.8M documents from
  Wikipedia.  After processing, it contains 482M observed words from a
  vocabulary of 7,700 terms.
\end{itemize}
For each collection, we set aside a test set of $10,000$ documents for
evaluating model fitness; these test sets were not given to the
algorithms for training.

\myparagraph{Evaluating model fitness.}  We evaluate how well a model
fits the data with the predictive distribution~\citep{Geisser:1975a}.
We are given a corpus and estimate its topics.  We then are given part
of a test document, which we use to estimate that document's topic
proportions.  Combining those topic proportions with the topics, we
form a predictive distribution over the vocabulary.  Under this
predictive distribution, a better model will assign higher probability
to the held-out words.

In more detail, we divide each test document's words $w$ into a set of
observed words $w_{\textrm{obs}}$ and held-out words
$w_{\textrm{ho}}$, keeping the sets of unique words in
$w_{\textrm{obs}}$ and $w_{\textrm{ho}}$ disjoint.  We approximate the
posterior distribution of topics $\beta$ implied by the training data
${\cal D}$, and then use that approximate posterior to estimate the
predictive distribution $p(w_{\textrm{new}} \g w_{\textrm{obs}}, {\cal
  D})$ of a new held-out word $w_{\textrm{new}}$ from the test
document.  Finally, we evaluate the log probability of the words in
$w_{\textrm{ho}}$ under this distribution.

This metric was used in~\cite{Teh:2007b} and \cite{Asuncion:2009}.
Unlike previous methods, like held-out perplexity~\citep{Blei:2003b},
evaluating the predictive distribution avoids comparing bounds or
forming approximations of the evaluation metric. It rewards a good
predictive distribution, however it is computed.

Operationally, we use the training data to compute variational
Dirichlet parameters for the topics.  We then use these parameters
with the observed test words $w_{\textrm{obs}}$ to compute the
variational distribution of the topic proportions.  Taking the inner
product of the expected topics and the expected topic proportions
gives the predictive distribution.

To see this is a valid approximation, note the following for a
$K$-topic LDA model,
\begin{eqnarray*}
  p(w_{\textrm{new}} \g {\cal D}, w_{\textrm{obs}}) &=&
  \int \int \left( 
  \textstyle \sum_{k=1}^{K}\theta_k \beta_{k,w_{\textrm{new}}}
  \right) p(\theta \g
  w_{\textrm{obs}}, \beta) p(\beta \g {\cal D}) d\theta d\beta \\
  &\approx& \int \int \left( \textstyle
    \sum_{k=1}^{K} \theta_k
    \beta_{kw_{\textrm{new}}} \right) q(\theta) q(\beta)d\theta d\beta \\
  &=& \textstyle\sum_{k=1}^K \E_q[\theta_k]
    \E_q[\beta_{k, w_{\textrm{new}}}],
%%   &=& \textstyle\sum_{k=1}^K \E_q[\theta_k \g w_{\textrm{obs}}, {\cal D}]
%%     \E_q[\beta_{k, w_{\textrm{new}}} \g {\cal D}].
\end{eqnarray*}
%% MDH: I got rid of the conditioning because I wasn't sure how to
%% make it clear that q(\theta) depends on q(\beta), but q(\beta)
%% doesn't depend on w_{obs}.
where $q(\beta)$ depends on the training data $\cal D$ and $q(\theta)$
depends on $q(\beta)$ and $w_{\textrm{obs}}$.
%% We have conditioned the expectations to make clear with which data the
%% $q$ distribution is fit.  
The metric independently evaluates each
held out word under this distribution.  In the HDP, the reasoning is
identical.  The differences are that the topic proportions are
computed via the two-level variational stick-breaking distribution and
$K$ is the truncation level of the approximate posterior.

\myparagraph{Setting the learning rate.}  Stochastic variational
inference introduces several parameters in setting the learning rate
schedule (see \myeq{stepsize}).  The forgetting rate $\kappa \in (0.5,
1]$ controls how quickly old information is forgotten; the delay $\tau
\ge 0$ down-weights early iterations; and the mini-batch size $S$ is
how many documents are subsampled and analyzed in each iteration.
Although stochastic variational inference algorithm converges to a
stationary point for any valid $\kappa$, $\tau$, and $S$, the quality
of this stationary point and the speed of convergence may depend on
how these parameters are set.

We set $\tau = 1$ and explored the following forgetting rates and
minibatch sizes:\footnote{We also explored various values of the delay
  $\tau$, but found that the algorithms were not sensitive.  To make
  this presentaton simpler, we fixed $\tau = 1$ in our report of the
  empirical study.}
\begin{itemize}
\item Forgetting rate $\kappa \in \{0.5, 0.6, 0.7, 0.8, 0.9, 1.0\}$
% \item Delay $\tau \in \{1, 10, 100, 1000\}$
\item Minibatch size $S\in\{10, 50, 100, 500, 1000\}$
\end{itemize}
We periodically paused each run to compute predictive likelihoods from
the test data.

\myparagraph{Results on LDA and HDP topic models.}  We studied LDA and
the HDP.  In LDA, we varied the number of topics $K$ to be 25, 50,
100, 200 and 300; we set the Dirichlet hyperparameters
$\alpha=1/K$. In the HDP, we set both concentration parameters
$\gamma$ and $\alpha$ equal to 1; we set the top-level truncation
$K=300$ and the second level truncation $T=20$. (Here $T \ll K $
because we do not expect documents to exhibit very many unique
topics.)  In both models, we set the topic Dirichlet parameter $\eta =
0.01$.  \myfig{nyt-topics} shows example topics from the HDP (on
\textit{New York Times} and \textit{Nature}).

\myfig{results} gives the average predictive log likelihood for both
models.  We report the value for a forgetting rate $\kappa=0.9$ and a
batch size of 500.  Stochastic inference lets us perform a large-scale
comparison of these models. The HDP gives consistently better
performance.  For larger numbers of topics, LDA overfits the data.  As
the modeling assumptions promise, the HDP stays robust to
overfitting.\footnote{Though not illustrated, we note that using the
  traditional measure of fit, held-out perplexity, does \textit{not}
  reveal this overfitting (though the HDP still outperforms LDA with
  that metric as well).  We feel that the predictive distribution is a
  better metric for model fitness.} That the HDP outperforms LDA
regardless of how many topics LDA uses may be due in part to the
additional modeling flexibility given by the corpus breaking
proportions $v$; these variables give the HDP the ability to say that
certain topics are a priori more likely to appear than others, whereas
the exchangeable Dirichlet prior used in LDA assumes that all topics
are equally common.

\begin{figure}
\begin{center}
  \begin{tabular}{l|c|c|c}
    & \textit{Nature} & \textit{New York Times} & \textit{Wikipedia} \\
    \hline
    LDA 25 & -7.24 & -7.73 & -7.44 \\
    LDA 50 & -7.23 & -7.68 & -7.43 \\
    LDA 100 & -7.26 & -7.66 & -7.41 \\
    LDA 200 & -7.50 & -7.78 & -7.64 \\
    LDA 300 & -7.86 & -7.98 & -7.74 \\
    HDP & \textbf{-6.97} & \textbf{-7.38} & \textbf{-7.07}
  \end{tabular}
\end{center}
\caption{\label{fig:results}Stochastic inference lets us compare
  performance on several large data sets.  We fixed the forgetting
  rate $\kappa = 0.9$ and the batch size to 500 documents.  We find
  that LDA is sensitive to the number of topics; the HDP gives
  consistently better predictive performance.  Traditional variational
  inference (on subsets of each corpus) did not perform as well as
  stochastic inference.}
\end{figure}

\begin{figure}
  \centerline{\includegraphics{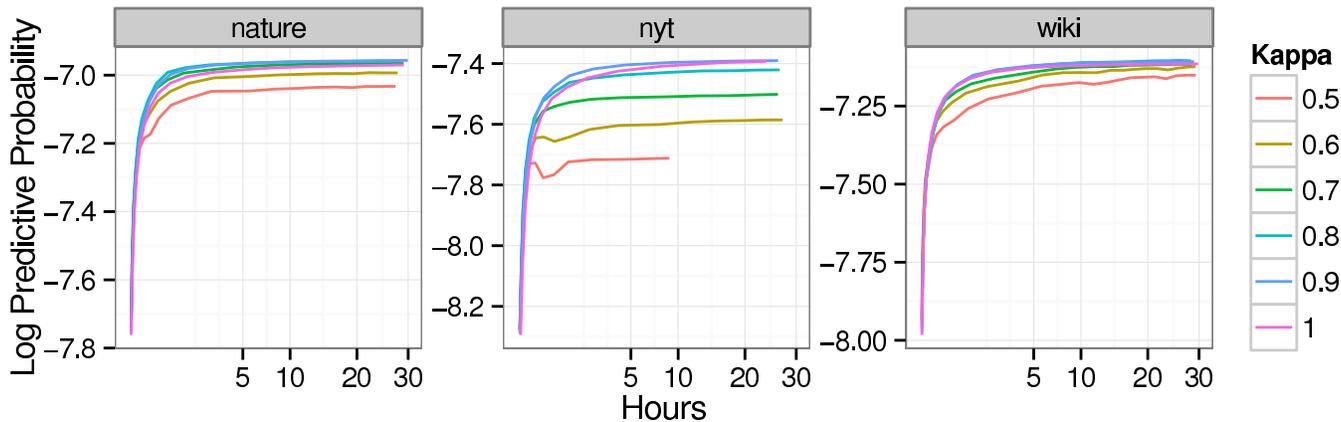}}
  \caption{\label{fig:hdp-kappa}HDP inference: Holding the batch size
    fixed at 500, we varied the forgetting rate $\kappa$.  Slower
    forgetting rates are preferred.}
\end{figure}

\begin{figure}
  \centerline{\includegraphics{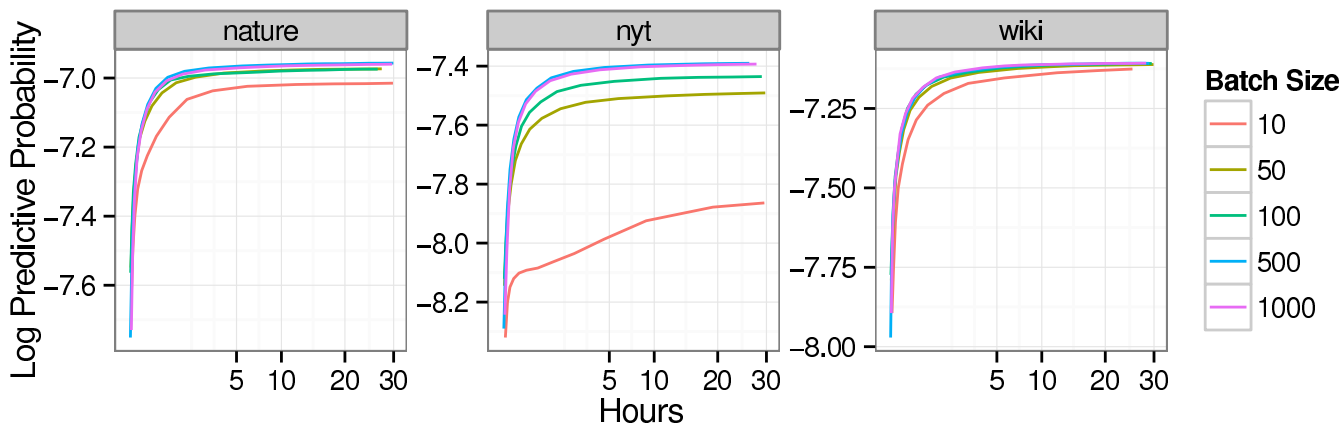}}
  \caption{\label{fig:hdp-batch}HDP inference: Holding the forgetting
    rate $\kappa$ fixed at 0.9, we varied the batch size. Batch sizes
    may be set too small (e.g., ten documents) but the difference in
    performance is small once set high enough.}
\end{figure}

We now turn to the sensitivity of stochastic inference to its learning
parameters.  First, we consider the HDP (the algorithm presented in
\myfig{stoch-hdp}).  We fixed the batch size to 500 and explored the
forgetting rate.\footnote{We fit distributions using the entire grid
  of parameters described above.  However, to simplify presenting
  results we will hold one of the parameters fixed and vary the
  other.}  \myfig{hdp-kappa} shows the results on all three corpora.
All three fits were sensitive to the forgetting rate; we see that a
higher value (i.e., close to one) leads to convergence to a better
local optimum.

Fixing the forgetting rate to 0.9, we explored various mini-batch
sizes.  \myfig{hdp-batch} shows the results on all three corpora.
Batch sizes that are too small (e.g., ten documents) can affect
performance; larger batch sizes are preferred.  That said, there was
not a big difference between batch sizes of 500 and 1,000.  The
\textit{New York Times} corpus was most sensitive to batch size; the
\textit{Wikipedia} corpus was least sensitive.

\begin{figure}[t]
  \centerline{\includegraphics{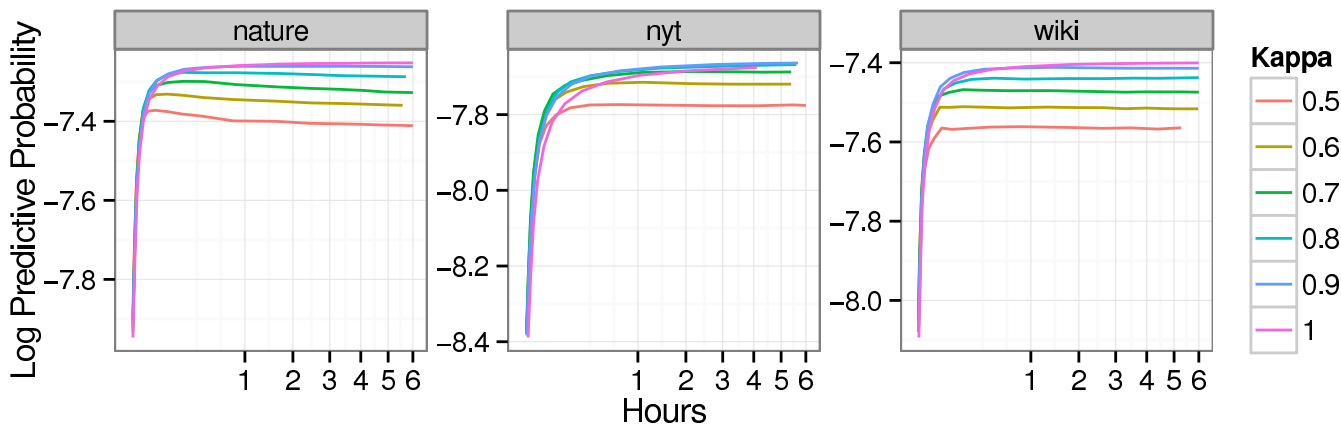}}
  \caption{\label{fig:lda-kappa}100-topic LDA inference: Holding the batch size
    fixed at 500, we varied the forgetting rate $\kappa$.  Slower
    forgetting rates are preferred.}
\end{figure}

\begin{figure}[t]
  \centerline{\includegraphics{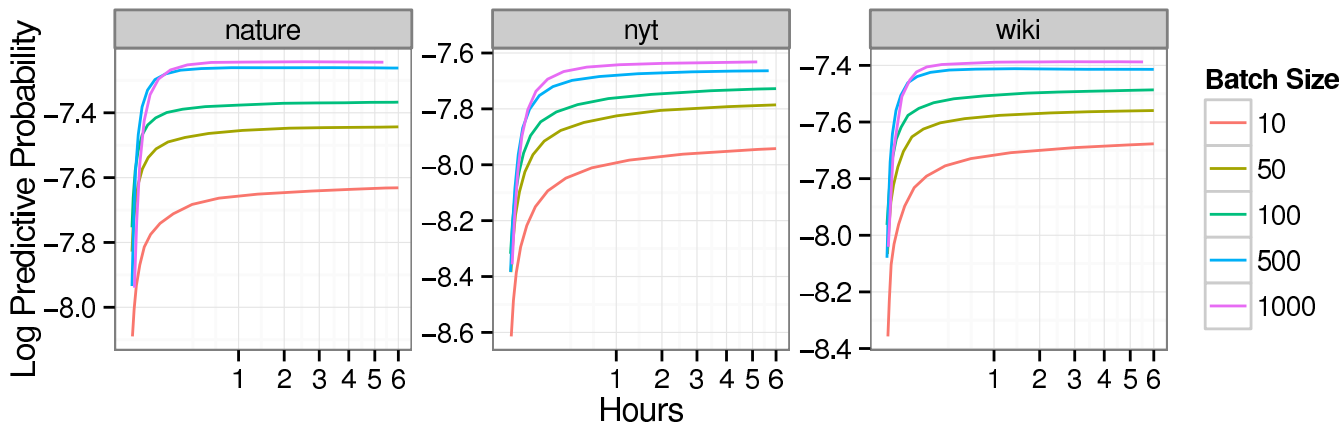}}
  \caption{\label{fig:lda-batch}100-topic LDA inference: Holding the
    learning rate $\kappa$ fixed at 0.9, we varied the batch size.
    Bigger batch sizes are preferred.}
\end{figure}

\myfig{lda-kappa} and \myfig{lda-batch} illustrate LDA's sensitivity
to the forgetting rate and batch size, respectively.  Again, we find
that large learning rates and batch sizes perform well.

\section{Discussion \label{sec:discussion}}

We developed stochastic variational inference, a scalable variational
inference algorithm that lets us analyze massive data sets with
complex probabilistic models.  The main idea is to use stochastic
optimization to optimize the variational objective, following noisy
estimates of the natural gradient where the noise arises by repeatedly
subsampling the data.  We illustrated this approach with two
probabilistic topic models, latent Dirichlet allocation and the
hierarchical Dirichlet process topic model.  With stochastic
variational inference, we can easily apply topic modeling to
collections of millions of documents.  More importantly, this
algorithm generalizes to many settings.

Since developing this algorithm, we have improved on stochastic
inference in a number of ways.  In~\cite{Gopalan:2012}, we applied it
to the mixed-membership stochastic blockmodel for uncovering
overlapping communities in large social networks.  This required
sampling non-uniformly from the data and adjusting the noisy gradient
accordingly.  In~\cite{Mimno:2012}, we developed a variant of
stochastic inference that combines MCMC for the local updates with
stochastic optimization for the global updates.  In topic modeling
this allows for efficient and sparse updates.  Finally,
in~\cite{Ranganath:2013}, we developed adaptive learning rates for
stochastic inference.  These outperform preset learning-rate schedules
and require less hand-tuning by the user.

Stochastic variational inference opens the door to several promising
research directions.

We developed our algorithm with conjugate exponential family models.
This class of models is expressive, but nonconjugate models---models
where a richer prior is used at the expense of mathematical
convenience---have expanded the suite of probabilistic tools at our
disposal.  For example, nonconjugate models can capture correlations
between topics~\citep{Blei:2007} or topics changing over
time~\citep{Blei:2006a,Wang:2008}, and the general algorithm presented
here cannot be used in these settings.  (In other work,
~\cite{Paisley:2012b} developed a stochastic variational inference
algorithm for a specific nonconjugate Bayesian nonparametric model.)
Recent research has developed general methods for non-conjugate
models~\citep{Knowles:2011,Gershman:2012a,Paisley:2012a,Wang:2013}.
Can these be scaled up with stochastic optimization?

We developed our algorithm with mean-field variational inference and
closed form coordinate updates.  Another promising direction is to use
stochastic optimization to scale up recent advances in variational
inference, moving beyond closed form updates and fully factorized
approximate posteriors.  As one example, collapsed variational
inference~\citep{Teh:2006,Teh:2007b} marginalizes out some of the
hidden variables, trading simple closed-form updates for a
lower-dimensional posterior.  As another example, structured
variational distributions relax the mean-field approximation, letting
us better approximate complex posteriors such as those arising in
time-series models~\citep{Ghahramani:1997,Blei:2006a}.

Finally, our algorithm lets us potentially connect innovations in
stochastic optimization to better methods for approximate posterior
inference. \cite{Wahabzada:2011} and \cite{Gopalan:2012} sample from
data non-uniformly to better focus on more informative data points.
We might also consider data whose distribution changes over time, such
as when we want to model an infinite stream of data but to ``forget''
data from the far past in a current estimate of the model. Or we can
study and try to improve our estimates of the gradient.  Are there
ways to reduce its variance, but maintain its unbiasedness?

\section*{Acknowledgments} David M. Blei is supported by NSF CAREER
IIS-0745520, NSF BIGDATA IIS-1247664, NSF NEURO IIS-1009542, ONR
N00014-11-1-0651, and the Alfred P. Sloan Foundation.  The authors are
grateful to John Duchi, Sean Gerrish, Lauren Hannah, Neil Lawrence,
Jon McAuliffe, and Rajesh Ranganath for comments and discussions.

In \mysec{inference}, we assumed that we can calculate $p(\beta | x,
z)$, the conditional distribution of the global hidden variables
$\beta$ given the local hidden variables $z$ and observed variables
$x$. In this appendix, we show how to do stochastic variational
inference under the weaker assumption that we can break the global
parameter vector $\beta$ into a set of $K$ subvectors $\beta_{1:K}$
such that each conditional distribution $p(\beta_k | x, z,
\beta_{-k})$ is in a tractable exponential family:
\begin{equation*}
\begin{split}
p(\beta_k | x, z, \beta_{-k}) =
h(\beta_k)\exp\{ \eta_g(x, z, \beta_{-k}, \alpha)^\top t(\beta_k)
-a_g(\eta_g(x, z, \beta_{-k}, \alpha)) \}.
\end{split}
\end{equation*}

We will assign each $\beta_k$ an independent variational distribution so
that
\begin{equation*}
\begin{split}
\textstyle
q(z, \beta) = (\prod_{n,j} q(z_{n,j})) \prod_k q(\beta_k).
\end{split}
\end{equation*}
We choose each $q(\beta_k)$ to be in the same exponential family as
the complete conditional $p(\beta_k | x, z, \beta_{-k})$,
\begin{equation*}
\begin{split}
q(\beta_k) = h(\beta_k) \exp\{ \lambda_k^\top t(\beta_k) - a_g(\lambda_k)\}.
\end{split}
\end{equation*}
We overload $h(\cdot)$, $t(\cdot)$, and $a(\cdot)$ so that, for
example, $q(\beta_k)=p(\beta_k|x, z, \beta_{-k})$ when
$\lambda_k=\eta_g(x, z, \beta_{-k}, \alpha)$.

The natural parameter $\eta_g(x, z, \beta_{-k}, \alpha)$ decomposes
into two terms,
\begin{equation*}
\begin{split}
\eta_g(x, z, \beta_{-k}, \alpha) =
\eta_g(\beta_{-k}, \alpha) + \sum_n \eta_g(x_n, z_n, \beta_{-k}, \alpha).
\end{split}
\end{equation*}
The first depends only on the global parameters $\beta_{-k}$ and the
hyperparameters $\alpha$; the second is a sum of $N$ terms that depend
on $\beta_{-k}$, $\alpha$, and a single local context $(x_n, z_n)$.

Proceeding as in \mysec{inference}, we will derive the natural
gradient of the ELBO implied by this model and choice of variational
distribution. Focusing on a particular $\beta_k$, we can write the
ELBO as
\begin{equation*}
\begin{split}
\cL &= \Eq[\log p(\beta_k | x, z, \beta_{-k})] - \Eq[\log q(\beta_k)]
+ \mathrm{const.}
\\ &= (\Eq[\eta_g(x, z, \beta_{-k}, \alpha)] - \lambda_k)^\top
\nabla_{\lambda_k} a_g(\lambda_k) + a_g(\lambda_k) + \mathrm{const.}
\end{split}
\end{equation*}
The gradient of $\cL$ with respect to $\lambda_k$ is then
\begin{equation*}
\begin{split}
\nabla_{\lambda_k}\cL =
\nabla^2_{\lambda_k} a_g(\lambda_k)
(\Eq[\eta_g(x, z, \beta_{-k}, \alpha)] - \lambda_k),
\end{split}
\end{equation*}
and the natural gradient of $\cL$ with respect to $\lambda_k$ is
\begin{equation*}
\begin{split}
\hat\nabla_{\lambda_k}\cL &= \Eq[\eta_g(x, z, \beta_{-k}, \alpha)] - \lambda_k
\\ &= -\lambda_k + 
\Eq[\eta_g(\beta_{-k}, \alpha)] +
\sum_n \Eq[\eta_g(x_n, z_n, \beta_{-k}, \alpha)].
\end{split}
\end{equation*}
Randomly sampling a local context $(x_i, z_i)$ yields a noisy (but
unbiased) estimate of the natural gradient,
\begin{equation*}
\begin{split}
\hat\nabla_{\lambda_k}\cL_i = -\lambda_k + 
\Eq[\eta_g(\beta_{-k}, \alpha)] +
N \Eq[\eta_g(x_i, z_i, \beta_{-k}, \alpha)]
\equiv
-\lambda_k + \hat\lambda_k.
\end{split}
\end{equation*}
We can use this noisy natural gradient exactly as in
\mysec{inference}. For each update $t$, we sample a context $(x_t,
z_t)$, optimize the local variational parameters $\phi_t$ by
repeatedly applying equation \myeq{coord-local}, and take a step of
size $\rho_t = (t + \tau)^{-\kappa}$ in the direction of the noisy
natural gradient:
\begin{equation}
\begin{split}
\label{eq:svi-lambdak}
\lambda_k^{(t)} = (1-\rho_t)\lambda_k^{(t-1)}
+ \rho_t\hat\lambda_k
\end{split}
\end{equation}
Note that the update in \myeq{svi-lambdak} depends only on
$\lambda^{(t-1)}$; we compute all elements of $\lambda^{(t)}$
simultaneously, whereas in a batch coordinate ascent algorithm
$\lambda_k^{(t)}$ could depend on $\lambda_{1:k-1}^{(t)}$.

\bibliography{bib}

\end{document}